\documentclass[letterpaper]{article} 
\usepackage[preprint]{aaai2027}  
\usepackage[hyphens]{url}  
\usepackage{graphicx} 
\usepackage{natbib}  
\usepackage{caption} 
\usepackage{amsmath}
\usepackage{amsfonts}
\usepackage{subcaption}
\usepackage{booktabs}
\usepackage{array}
\usepackage{arydshln}

\usepackage[colorlinks=true,allcolors={[rgb]{0.10,0.45,0.80}}]{hyperref}
\makeatletter
\global\expandafter\let\csname ver@hyperref.sty\endcsname\relax
\makeatother

\title{HeadCast: Casting Attention Heads for Efficient Autoregressive Video Generation}

\author{
    Jinliang Shen\textsuperscript{\rm 1}\thanks{\ Work done during an internship at KlingAI Research.},
    Lianghao Su\textsuperscript{\rm 2},
    Zheming Li\textsuperscript{\rm 2},
    Kang He\textsuperscript{\rm 2},
    Ziliang Lai\textsuperscript{\rm 2},
    Yanbing Jiang\textsuperscript{\rm 1}\thanks{\ Corresponding authors.},
    Chengru Song\textsuperscript{\rm 2}\footnotemark[2]
}
\affiliations{
    \textsuperscript{\rm 1}Peking University\\
    \textsuperscript{\rm 2}KlingAI Research\\
    sjl@stu.pku.edu.cn, jyb@ss.pku.edu.cn, songchengru@klingai.com
}

\newlength{\hw}
\newcommand{\headrows}[1]{%
  \includegraphics[width=\hw]{Figures/appendix/#1_f6_raw.pdf} &
  \includegraphics[width=\hw]{Figures/appendix/#1_f6_frame.pdf} &
  \includegraphics[width=\hw]{Figures/appendix/#1_f6_spatial.pdf} \\
  \includegraphics[width=\hw]{Figures/appendix/#1_f18_raw.pdf} &
  \includegraphics[width=\hw]{Figures/appendix/#1_f18_frame.pdf} &
  \includegraphics[width=\hw]{Figures/appendix/#1_f18_spatial.pdf} \\[5pt]}

\begin{document}

\maketitle

\pagestyle{plain}
\thispagestyle{plain}

\begin{abstract}
Autoregressive (AR) video diffusion models have become a promising paradigm for long and streaming video synthesis, but the continuously growing Key-Value (KV) cache makes attention the dominant inference cost, especially at high resolution where each frame contributes many tokens. Existing remedies either evict the cache with coarse heuristics that cause inter-frame flickering, or require model re-training. We propose \textbf{HeadCast}, a training-free, plug-and-play acceleration framework built on the observation that a pre-trained AR model's attention heads exhibit stable, heterogeneous behaviors. After a short warm-up, HeadCast performs a one-time classification at the maximum-noise step that sorts every head into one of four archetypes---\textbf{Sink}, \textbf{Dummy}, \textbf{Spatial}, and \textbf{Global}---and restructures the monolithic KV cache into head-specific pathways. Crucially, it retains the Global heads that preserve the long-range temporal consistency aggressive eviction destroys. Because the Spatial pathway operates on a fixed-size grid, its savings grow with resolution: across state-of-the-art AR models, HeadCast accelerates inference by up to \textbf{1.62$\times$} at 720P and \textbf{1.95$\times$} at 1080P, while keeping VBench quality on par with full attention and largely flicker-free. Code: \url{https://github.com/sjlgaga/HeadCast}.
\end{abstract}

\section{Introduction}

Diffusion Transformers~\cite{peebles2023dit} have advanced high-fidelity video generation~\cite{ho2022vdm,blattmann2023svd,yang2024cogvideox,kong2024hunyuan,wan2025}. To avoid generating all frames at once, autoregressive (AR) models synthesize video block-by-block, supporting long-horizon and streaming generation and reusing historical states through Key-Value (KV) caching.

As generation proceeds, however, the KV cache grows without bound, and attention over it incurs an $\mathcal{O}(L^2)$ cost that comes to dominate inference. A sliding window bounds this growth, but for high-resolution video the in-window sequence is still long---each frame contributes many tokens---so inference remains slow. Aggressively shrinking the cache is no remedy either: Dummy Forcing~\cite{guo2026dummyhead} relies on a coarse head classification and evicts long-range context that some heads depend on, producing inter-frame flickering and structural drift, while training-based sparse-attention methods require costly re-training.

This raises a question: must every attention head attend to the full historical context, or do heads differ in what they need? Visualizing pre-trained AR video models, we find that attention heads exhibit stable, heterogeneous structural preferences. We identify four archetypes: \textbf{Sink} heads that anchor on the initial frame, \textbf{Dummy} heads that attend only to the most recent block, \textbf{Spatial} heads that attend to a local spatial neighborhood across history, and \textbf{Global} heads that require the full context. These assignments remain stable throughout generation, so a single early classification suffices.

Building on this, we propose \textbf{HeadCast}, a training-free, plug-and-play acceleration framework that \emph{casts} each pre-trained attention head to a dedicated computation path. After a brief full-context warm-up, HeadCast runs a one-time classification at the maximum-noise step ($t=1000$), where attention reflects structural rather than content-specific preferences, and routes each head to a tailored pathway: Sink and Dummy heads keep a single block, Spatial heads attend within a fixed grid, and Global heads retain the full sliding window. The monolithic KV cache is correspondingly split into compact, head-specific buffers. Retaining the Global heads preserves temporal consistency and avoids the flickering from over-aggressive eviction.

Because the Spatial pathway is confined to a fixed-size grid, HeadCast's speedup grows with the KV-cache size, reaching up to $1.62\times$ at 720P and $1.95\times$ at 1080P without any training. Across Self-Forcing~\cite{huang2025selfforcing}, LongLive~\cite{yang2026longlive}, Causal Forcing~\cite{zhu2026causalforcing}, and Reward Forcing~\cite{lu2026rewardforcing}, it preserves VBench quality and frame-level fidelity (PSNR/LPIPS) while removing the flickering that aggressive eviction induces.

In summary, our main contributions are as follows:
\begin{itemize}
    \item We identify and categorize four stable, heterogeneous attention-head archetypes---Sink, Dummy, Spatial, and Global---in pre-trained AR video diffusion models.
    \item We propose HeadCast, a training-free framework that classifies heads once and restructures the KV cache into head-specific computation pathways, explicitly retaining the Global heads for temporal consistency.
    \item Across multiple AR models, HeadCast delivers acceleration that scales with resolution (up to $1.62\times$ at 720P, $1.95\times$ at 1080P) while preserving visual fidelity and temporal coherence.
\end{itemize}

\section{Related Work}
\subsection{Autoregressive Video Diffusion}

Diffusion models~\cite{ho2020ddpm,song2021ddim,rombach2022ldm} have become the dominant paradigm for high-fidelity visual synthesis, and a growing line of work~\cite{chen2024diffusionforcing,yin2025causvid,huang2025selfforcing,liu2025rolling,yang2026longlive,sandai2025magi} combines diffusion modeling with autoregressive (AR) prediction to support long-horizon, streaming video generation, reducing cost through causal modeling and Key-Value (KV) caching. MAGI-1~\cite{sandai2025magi} generates videos chunk-by-chunk with progressive per-chunk denoising, enabling streaming synthesis. CausVid~\cite{yin2025causvid} converts a pre-trained bidirectional diffusion transformer into a causal AR generator with KV caching, and Self-Forcing~\cite{huang2025selfforcing} addresses the train-inference mismatch by conditioning the model on its own generated frames. Building on these ideas, Rolling Forcing~\cite{liu2025rolling} expands the diffusion window to suppress error accumulation, while LongLive~\cite{yang2026longlive} introduces KV re-caching to maintain visual continuity across scene transitions. Many of these AR generators are additionally compressed into few-step samplers through diffusion distillation~\cite{song2023consistency,yin2024dmd,salimans2022progressive}, which cuts the \emph{number} of denoising steps; HeadCast is orthogonal and fully training-free, instead reducing the per-step attention cost by exploiting the heterogeneous attention behaviors these pre-trained models already exhibit.

\subsection{KV Cache Compression}

The attention mechanism~\cite{vaswani2017attention} underlies these models, yet its linearly growing memory footprint has motivated extensive KV cache compression, primarily for Large Language Models (LLMs). Token-level methods such as StreamingLLM~\cite{xiao2023streamingllm} preserve initial-token ``sinks,'' while H2O~\cite{zhang2023h2o} and SnapKV~\cite{li2024snapkv} keep pivotal tokens by their attention scores; head-specific methods such as DuoAttention~\cite{xiao2024duoattention} and FastGen~\cite{ge2023fastgen} assign different cache budgets to different heads. Complementary directions instead attack the cost at the systems level through IO-aware kernels~\cite{dao2022flashattention} or shrink the cache via low-bit quantization~\cite{liu2024kivi}. These target the 1D structure of text and do not exploit the spatial-temporal redundancy of video. HeadCast extends head-level cache restructuring to AR video diffusion with dimension-aware, per-head-type policies.

\subsection{Efficient Video Generation}

Early video models adopted bidirectional Diffusion Transformers (DiT)~\cite{peebles2023dit}, accelerated mainly via fixed-length sparse attention~\cite{zhang2025sta, xia2025vsa}. The Sparse VideoGen family~\cite{xi2025svg,yang2025svg2} identifies spatial and temporal heads (and, in SVG2, clustered token blocks) for intra- and inter-frame sparsity. These patterns are defined under \emph{bidirectional} 3D full attention, however, and do not directly transfer to the causal, dynamically growing context of AR models.

Closer to our setting, recent methods accelerate AR video models directly. Light Forcing~\cite{lv2026lightforcing} allocates chunk-level sparsity with coarse-to-fine top-$k$ selection but applies one scheme to all heads, and Sparse Forcing~\cite{xu2026sparseforcing} learns native sparse attention---both requiring additional training. A training-free line instead compresses the KV cache directly for speed: Dummy Forcing~\cite{guo2026dummyhead} and the concurrent Forcing-KV~\cite{ji2026forcingkv}. The closest to ours, Dummy Forcing, forces many heads with only mild temporal sparsity into an aggressive ``dummy'' mode that keeps just the current frame, over-discarding the long-range context they depend on and---as our experiments show---lowering frame-level fidelity and inducing inter-frame flickering. (Deep Forcing~\cite{yi2026deepforcing}, also training-free, instead targets long-range generation quality and stability rather than resolution-scaling acceleration.) Concurrent to ours, Head Forcing~\cite{tian2026headforcing} likewise exploits attention-head heterogeneity, but for \emph{long-horizon extrapolation}---extending generation to minute length via a hierarchical memory and head-wise RoPE re-encoding---rather than the resolution-scaling acceleration we target. HeadCast is also training-free, yet adds a \emph{Spatial} type that captures these heads' spatial locality (recovering SVG's locality in the causal setting) and explicitly retains a \emph{Global} type for the rest, attaining substantially higher fidelity than Dummy Forcing at only a modest speed cost---a markedly better speed--fidelity trade-off.

\begin{figure*}[t]
    \centering
    \begin{subfigure}[b]{0.48\textwidth}
        \centering
        \includegraphics[width=\textwidth]{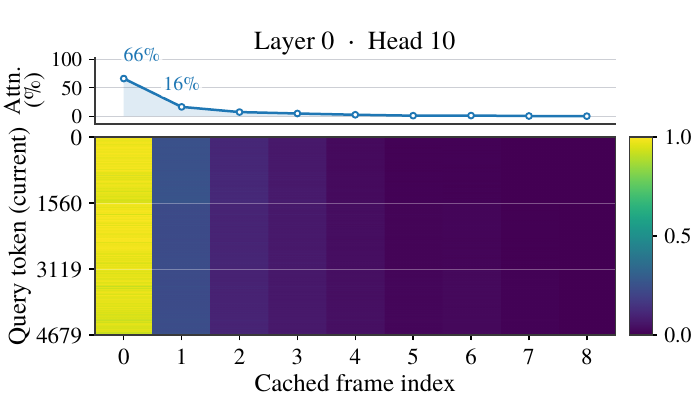}
        \caption{Sink}
        \label{fig:1a}
    \end{subfigure}
    \hfill
    \begin{subfigure}[b]{0.48\textwidth}
        \centering
        \includegraphics[width=\textwidth]{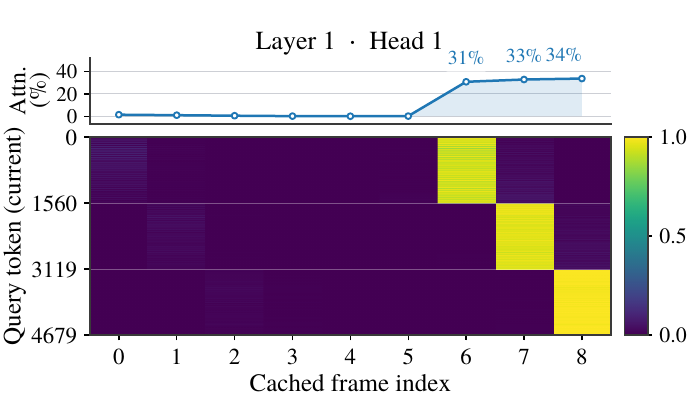}
        \caption{Dummy}
        \label{fig:1b}
    \end{subfigure}

    \begin{subfigure}[b]{0.48\textwidth}
        \centering
        \includegraphics[width=\textwidth]{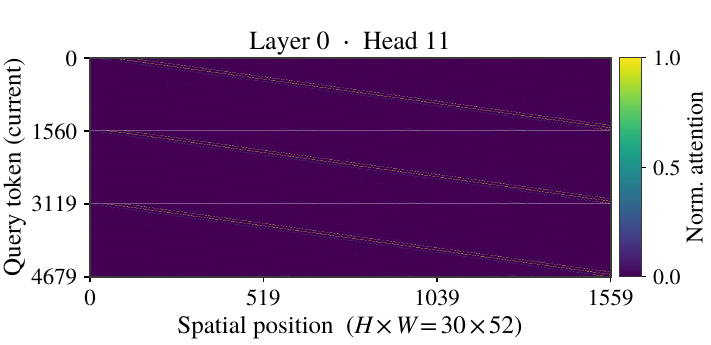}
        \caption{Spatial}
        \label{fig:1c}
    \end{subfigure}
    \hfill
    \begin{subfigure}[b]{0.48\textwidth}
        \centering
        \includegraphics[width=\textwidth]{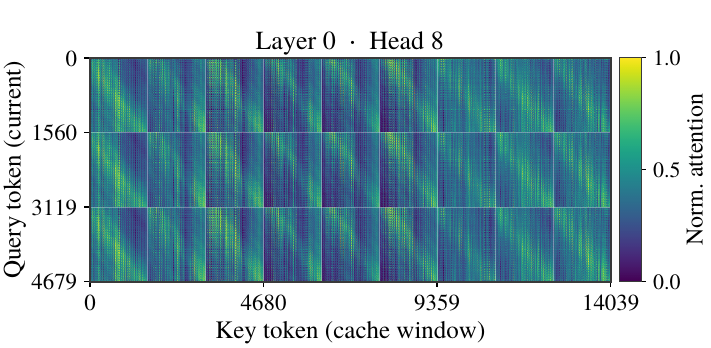}
        \caption{Global}
        \label{fig:1d}
    \end{subfigure}
    \caption{The four attention-head archetypes in a pre-trained AR video model. \textbf{(a)}~Sink and \textbf{(b)}~Dummy heads (per-frame view) concentrate on the first and the most recent block, respectively; \textbf{(c)}~a Spatial head (per-position view) attends to each query's local neighborhood; \textbf{(d)}~a Global head (raw map) spreads across the whole history.}
    \label{fig:horizontal_four}
\end{figure*}

\section{Motivation}

\subsection{Preliminary: Autoregressive Video Diffusion and KV Cache}

Autoregressive video diffusion models generate video block-by-block. Given a video of $T$ frames $X = \{x_1, x_2, \ldots, x_T\}$, the model learns the joint distribution under a causal factorization:
\begin{equation}
p(x_1, \ldots, x_T) = \prod_{i=1}^{T} p(x_i \mid x_{<i})
\end{equation}
Each conditional $p(x_i \mid x_{<i})$ is a denoising diffusion model~\cite{peebles2023dit} that denoises the current latents from Gaussian noise, conditioned on previously generated frames $x_{<i}$. Each autoregressive (AR) step generates a chunk of frames under a block-causal mask that blocks leakage from future frames, so the keys and values (KV) of historical frames can be cached and reused, avoiding recomputation.

However, the accumulated KV cache grows linearly with video length, inducing an $\mathcal{O}(L^2)$ attention cost that soon dominates inference. A sliding window of recent frames~\cite{huang2025selfforcing} curbs this cost but exposes a tension: at high resolution the in-window sequence is still long enough to keep inference slow, yet shrinking the window discards the long-range context longer videos need for temporal coherence. Retaining the first frame as a fixed sink stabilizes quality, hinting that not all cached content matters equally. Dummy Forcing~\cite{guo2026dummyhead} manages the cache per head, but evicts too coarsely, dropping context heads depend on and causing quality loss and flicker. Efficiency without quality loss thus demands a finer head classification aligned with each head's true temporal dependencies.

\begin{figure*}[t]
    \centering
    \includegraphics[width=0.95\textwidth]{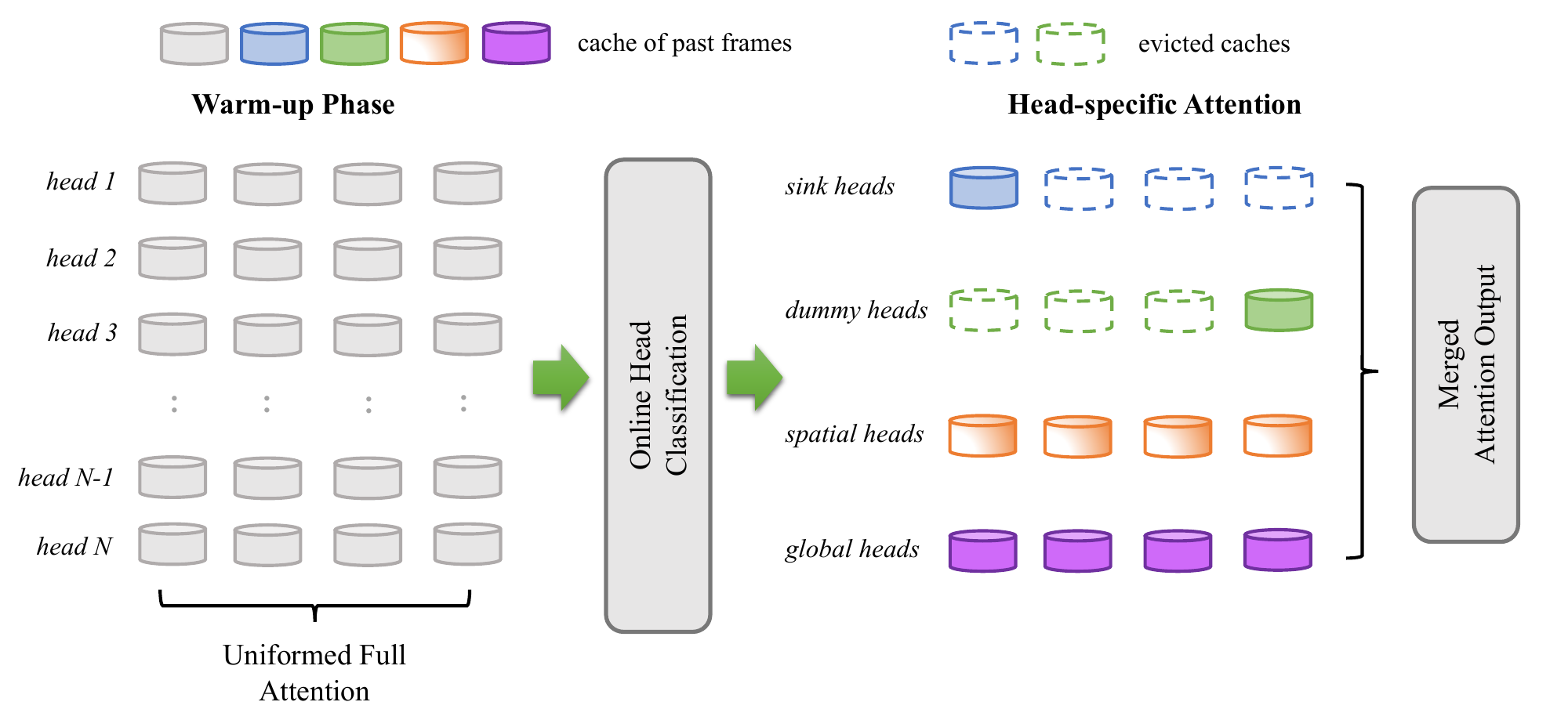}
    \caption{The four-phase HeadCast pipeline: full-context Warm-up, Online Classification, Heterogeneous Cache Management, and Head-Specific Attention.}
    \label{fig:headcast_overview}
\end{figure*}

\subsection{Heterogeneous Attention Patterns in Video DiT}

In current AR video diffusion models, every attention head attends over the entire accumulated history. We hypothesize that this is unnecessary---heads exhibit intrinsically diverse dependencies on the historical context---and visualize the attention maps of an autoregressive baseline (e.g., Self-Forcing) during inference to expose their heterogeneous patterns.

\noindent\textbf{\textit{Observation 1: Attention heads display sparsity in the temporal scope.}}

Inspired by Dummy Forcing~\cite{guo2026dummyhead}, we first investigate how different attention heads attend to historical frames. In Figure~\ref{fig:horizontal_four}(a)--(b), for each query token, we aggregate its attention scores across all key tokens within the same frame to quantify the importance of each cached frame. As illustrated in the figure, numerous attention heads exhibit significant sparsity along the temporal dimension, falling into two distinct patterns:

\noindent\textbf{Sink Pattern:} The majority of the attention weights from all query tokens strongly congregate on the first block of the KV cache.

\noindent\textbf{Dummy Pattern:} The attention weights of the query tokens are almost exclusively concentrated on the current denoising block.

\noindent\textbf{\textit{Observation 2: Attention heads display sparsity in the spatial scope.}}

For heads without extreme temporal sparsity, we probe the spatial dimension by projecting the historical KV cache onto the original 2D spatial grid (height and width) and aggregating the attention score at each spatial position, which reveals each query's spatial receptive field. As shown in Figure~\ref{fig:horizontal_four}(c), some heads concentrate the vast majority of a query's attention on key tokens within its immediate spatial neighborhood. We thus split the remaining heads into two further patterns:

\noindent\textbf{Spatial Pattern:} The attention weights of a query token strongly congregate on the key tokens located within its localized spatial neighborhood across historical frames.

\noindent\textbf{Global Pattern:} The attention weights spread across the entire KV cache, so these heads require the full historical context. Figure~\ref{fig:horizontal_four}(d) shows the raw attention score map of such a global head, whose attention is distributed globally along both the temporal and spatial dimensions.

\noindent\textbf{\textit{Observation 3: Different attention heads exhibit pattern stability over inference steps.}}

Crucially, most attention heads retain their archetype across autoregressive steps, denoising timesteps, and text prompts. Quantitatively, over $\mathbf{90\%}$ of the heads receive the \emph{identical} archetype across autoregressive steps (frame $12$ vs.\ $18$, two blocks apart), and the assignment stays similarly stable across denoising timesteps and prompts (full per-axis agreement rates appear in Figure~\ref{fig:stability_bar}). This stability lets us exploit the heterogeneous sparsity with a single early classification, at no recurring cost.

\section{HeadCast}

\subsection{Overview}

To exploit historical-context sparsity without sacrificing visual fidelity, we present HeadCast, a training-free inference framework that gives each attention head a tailored computation path and cache-management policy. As illustrated in Figure~\ref{fig:headcast_overview}, its pipeline has four phases. A brief \textit{Warm-up} runs full attention to accumulate a stable context; once the cache exceeds the sliding-window size $W$, a one-time \textit{Online Classification} sorts every head into one of four archetypes---Sink, Dummy, Spatial, or Global---from the cosine similarity between its full- and restricted-context outputs. \textit{Heterogeneous Cache Management} then splits the monolithic cache into type-specific buffers along the head dimension, and \textit{Head-Specific Attention} routes each head through its own path, jointly cutting sequence length and attention FLOPs. We detail each below.

\subsection{Online Classification}
To uncover each head's dependency profile without training, we introduce a one-time \textit{Online Classification}, triggered once the accumulated KV cache first exceeds the window size $W$. By then, the \textit{Warm-up} phase has accumulated a full historical context that serves as the classification reference.

\subsubsection{Classification Metrics and Archetypes}
Given the query tensor $Q$ and the full historical key-value pairs $(K_{\text{full}}, V_{\text{full}})$ at the classification step, we first compute the unconstrained reference output for each head $h$:
\begin{equation}
O_{\text{ref}}^{(h)} = \text{Attention}\left(Q^{(h)}, K_{\text{full}}^{(h)}, V_{\text{full}}^{(h)}\right)
\end{equation}
To probe how each head $h$ behaves under a constrained context, we define three restricted attention operations, each over a designated subset of the history:
\begin{itemize}
    \item \textbf{Sink Proxy:} Attention restricted to the first temporal block (the semantic anchor), yielding $O_{\text{sink}}^{(h)}$.
    \item \textbf{Dummy Proxy:} Attention restricted to the current local block of most recent frames, yielding $O_{\text{dummy}}^{(h)}$.
    \item \textbf{Spatial Proxy:} Attention over all historical blocks but confined to a local $(2r+1) \times (2r+1)$ grid centered on each query token, yielding $O_{\text{spatial}}^{(h)}$.
\end{itemize}

For each head $h$, we quantify the alignment between the restricted outputs and the reference by averaging the cosine similarity across all $L_q$ query tokens:
\begin{equation}
\begin{aligned}
\text{cos}_{m}^{(h)} &= \frac{1}{L_q} \sum_{t=1}^{L_q} \cos\!\left(O_{m}^{(h)}[t],\, O_{\text{ref}}^{(h)}[t]\right), \\
&\quad m \in \{\text{sink},\, \text{dummy},\, \text{spatial}\}.
\end{aligned}
\end{equation}
Because averaged cosine similarity can stay high even when a head fails at a few boundary tokens---which we find leaves residual ghosting if such heads are routed to the Spatial path---we adopt a conservative, worst-case score for the spatial case:
\begin{equation}
\text{score}^{(h)} = \mathcal{P}_5\left(\text{cos}_{\text{spatial}}^{(h)}\right) - \gamma \cdot \text{MSE}\left(O_{\text{spatial}}^{(h)}, O_{\text{ref}}^{(h)}\right)
\end{equation}
where $\mathcal{P}_5(\cdot)$ is the 5th percentile of the token-wise cosine similarities and $\gamma$ weights the MSE penalty.

\subsubsection{Mutually Exclusive Decision Rules}
To assign each head to a single mode, we apply a mutually exclusive decision hierarchy with empirical thresholds ($\theta_s, \theta_d, \theta_{\text{sp}}, \theta_{\text{sc}}$):
\begin{equation}
\mathcal{C}(h) = \begin{cases}
\text{Sink} & \text{if } \text{cos}_{\text{sink}}^{(h)} \geq \theta_s \\
\text{Dummy} & \text{elif } \text{cos}_{\text{dummy}}^{(h)} \geq \theta_d \\
\text{Spatial} & \text{elif } \text{cos}_{\text{spatial}}^{(h)} \geq \theta_{\text{sp}} \text{ and } \text{score}^{(h)} \geq \theta_{\text{sc}} \\
\text{Global} & \text{otherwise}
\end{cases}
\end{equation}

This classification runs only once, on a single autoregressive block, so it adds only a one-time overhead---under $1.5\%$ of generation time on $30$-second LongLive and $\sim$$5$--$8\%$ on $5$-second clips, with no steady-state cost (full breakdown in Appendix~\ref{sec:overhead}).

\begin{figure}[t]
    \centering
    \includegraphics[width=\columnwidth]{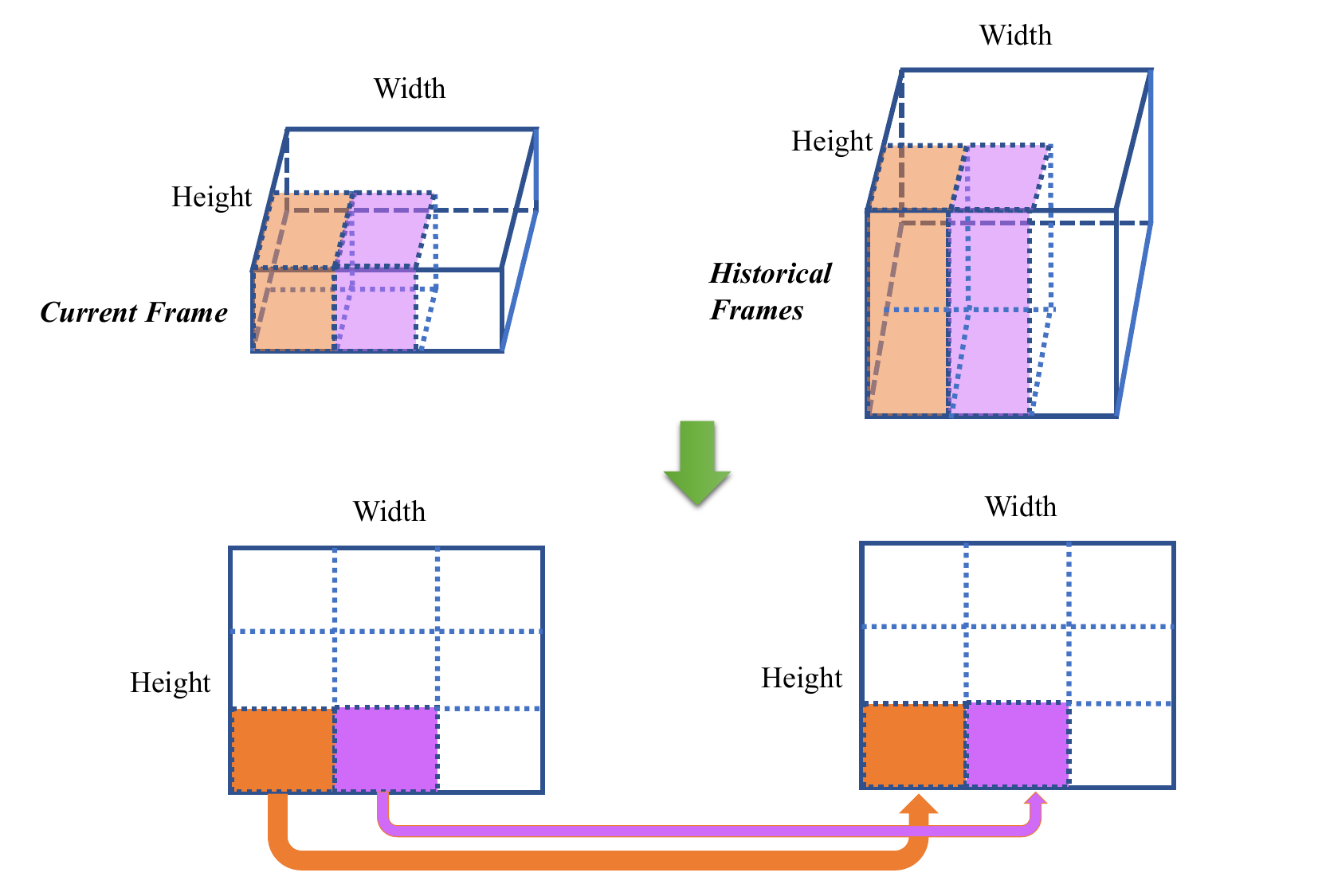}
    \caption{The Spatial Path. Each frame is partitioned into non-overlapping grids; a query attends only to KV tokens in its own grid across history, decomposing dense attention into independent sub-matrices.}
    \label{fig:spatial_head}
\end{figure}

\begin{figure*}[!t]
    \centering
    \footnotesize
    \setlength{\tabcolsep}{4pt}
    \begin{tabular}{l c c c c c c c c}
        \toprule
        \textbf{Method} & \textbf{FPS $\uparrow$} & \textbf{Speedup $\uparrow$} & \textbf{Quality $\uparrow$} & \textbf{Semantic $\uparrow$} & \textbf{Total $\uparrow$} & \textbf{Dynamic Degree} & \textbf{PSNR $\uparrow$} & \textbf{LPIPS $\downarrow$} \\
        \midrule
        $\diamond$ \textbf{Self-Forcing} (5s) & 19.48 & 1.00$\times$ & 84.60 & 80.77 & 83.84 & 64.72 & --- & --- \\
        \quad $\bullet$ + Dummy Forcing & 20.30 & 1.04$\times$ & 84.22 & 80.77 & 83.53 & 55.83 & 18.23 & 0.1919 \\
        \quad $\bullet$ \textbf{+ HeadCast} & 20.38 & \textbf{1.05$\times$} & \textbf{84.66} & 80.77 & \textbf{83.88} & 64.72 & \textbf{25.64} & \textbf{0.0469} \\
        \midrule
        $\diamond$ \textbf{LongLive} (30s) & 16.06 & 1.00$\times$ & 82.77 & 80.17 & 82.26 & 35.59 & --- & --- \\
        \quad $\bullet$ + Dummy Forcing & 18.63 & \textbf{1.16$\times$} & \textbf{82.93} & 80.69 & \textbf{82.48} & 34.81 & 10.27 & 0.6161 \\
        \quad $\bullet$ \textbf{+ HeadCast} & 17.29 & 1.08$\times$ & 82.42 & \textbf{80.93} & 82.12 & 29.84 & \textbf{11.02} & \textbf{0.5779} \\
        \midrule
        $\diamond$ \textbf{Causal Forcing} (5s) & 20.08 & 1.00$\times$ & 85.42 & 80.95 & 84.53 & 85.0 & --- & --- \\
        \quad $\bullet$ + Dummy Forcing & 20.37 & 1.01$\times$ & 84.58 & 80.86 & 83.83 & 94.6 & 15.96 & 0.2237 \\
        \quad $\bullet$ \textbf{+ HeadCast} & 20.46 & \textbf{1.02$\times$} & \textbf{85.37} & \textbf{81.01} & \textbf{84.50} & 95.2 & \textbf{22.42} & \textbf{0.0678} \\
        \midrule
        $\diamond$ \textbf{Reward Forcing} (5s) & 19.81 & 1.00$\times$ & 84.94 & 80.88 & 84.13 & 68.33 & --- & --- \\
        \quad $\bullet$ + Dummy Forcing & 20.17 & \textbf{1.02$\times$} & 84.64 & \textbf{80.90} & 83.89 & 58.05 & 18.63 & 0.1507 \\
        \quad $\bullet$ \textbf{+ HeadCast} & 20.14 & 1.02$\times$ & \textbf{84.79} & 80.87 & \textbf{84.00} & 65.56 & \textbf{24.92} & \textbf{0.0449} \\
        \bottomrule
    \end{tabular}
    \captionof{table}{Main results on four autoregressive video models at standard 480P resolution (clip length shown per backbone). We report VBench (Quality, Semantic, Total) and Dynamic Degree, with PSNR and LPIPS computed against each model's full-attention output (hence ``---'' for the baseline).}
    \label{tab:main_results}

    \vspace{10pt}
    \setlength{\tabcolsep}{1.5pt}
    \renewcommand{\arraystretch}{0.55}
    \begin{tabular}{@{}m{2.2cm}@{\hskip 3pt}ccccc@{}}
        & \footnotesize Frame 127 & \footnotesize Frame 128 & \footnotesize Frame 129 & \footnotesize Frame 130 & \footnotesize Frame 131 \\
        \footnotesize Dummy Forcing &
        \includegraphics[width=0.168\textwidth]{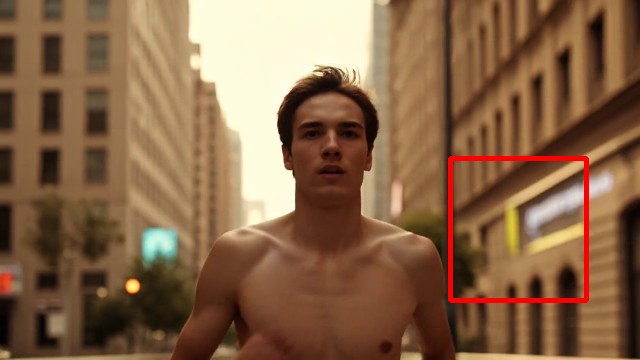} &
        \includegraphics[width=0.168\textwidth]{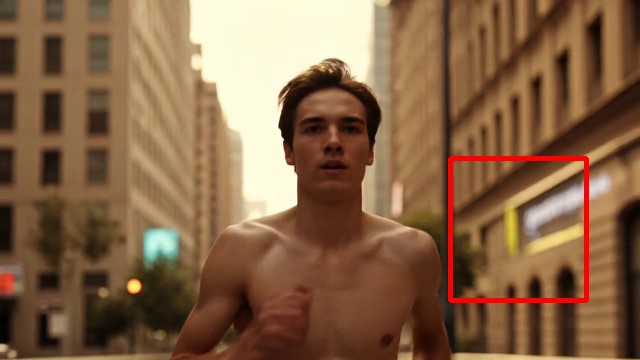} &
        \includegraphics[width=0.168\textwidth]{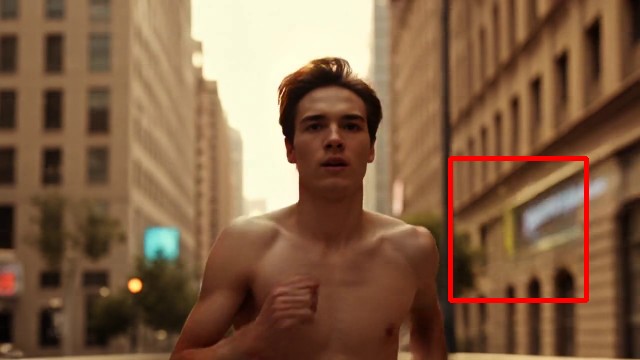} &
        \includegraphics[width=0.168\textwidth]{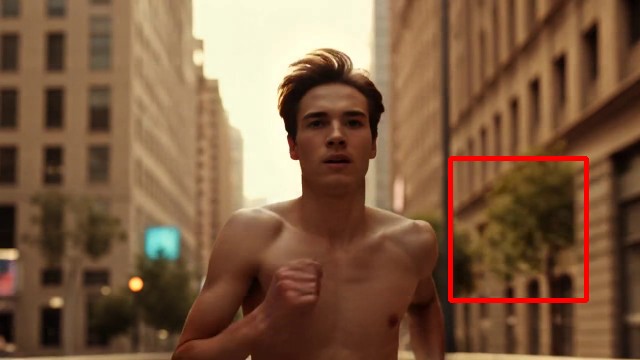} &
        \includegraphics[width=0.168\textwidth]{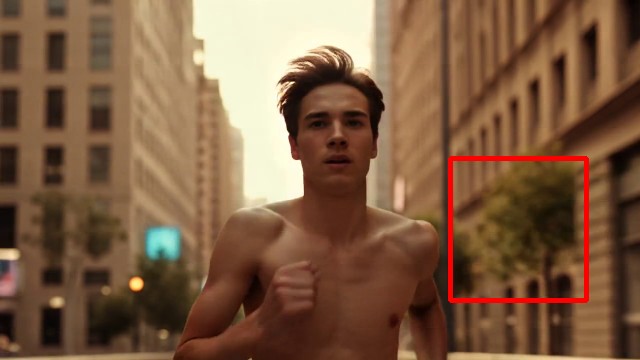} \\
        \footnotesize Full Attention &
        \includegraphics[width=0.168\textwidth]{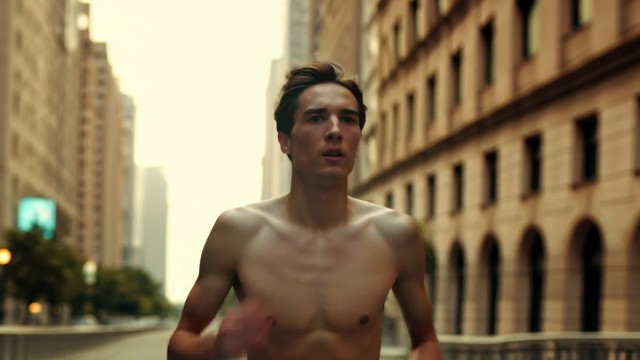} &
        \includegraphics[width=0.168\textwidth]{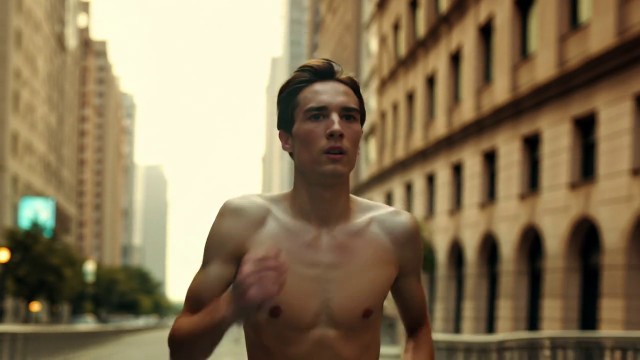} &
        \includegraphics[width=0.168\textwidth]{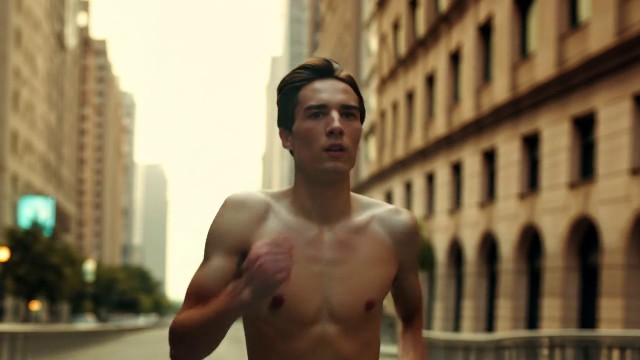} &
        \includegraphics[width=0.168\textwidth]{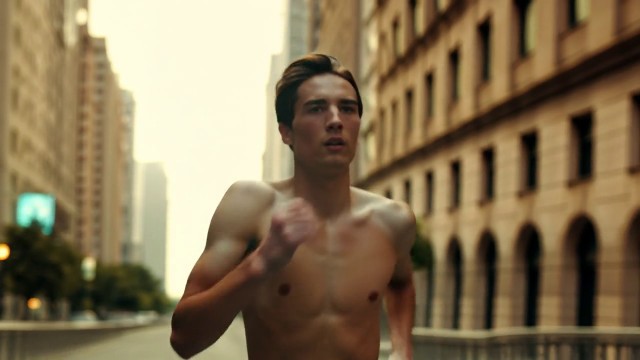} &
        \includegraphics[width=0.168\textwidth]{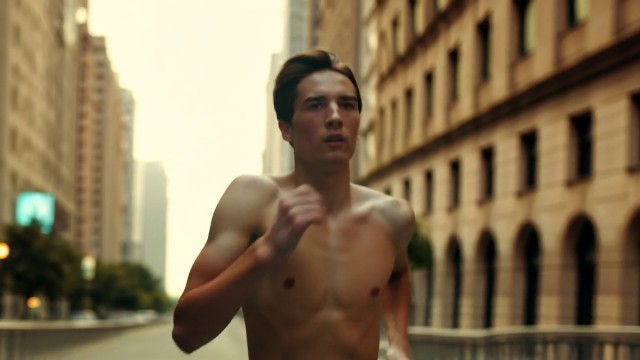} \\
        \footnotesize HeadCast &
        \includegraphics[width=0.168\textwidth]{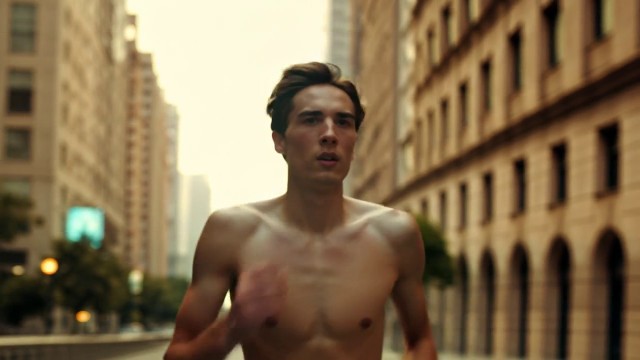} &
        \includegraphics[width=0.168\textwidth]{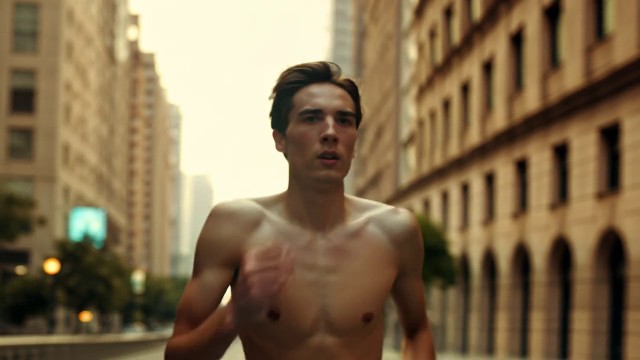} &
        \includegraphics[width=0.168\textwidth]{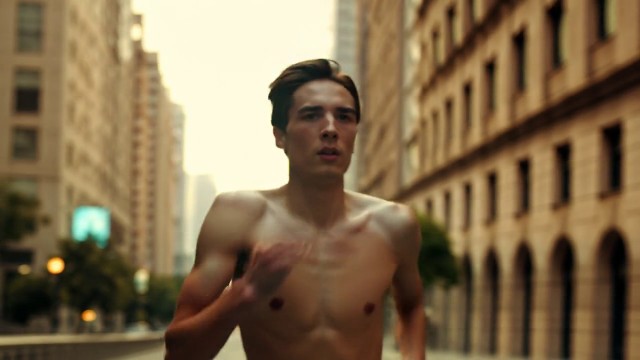} &
        \includegraphics[width=0.168\textwidth]{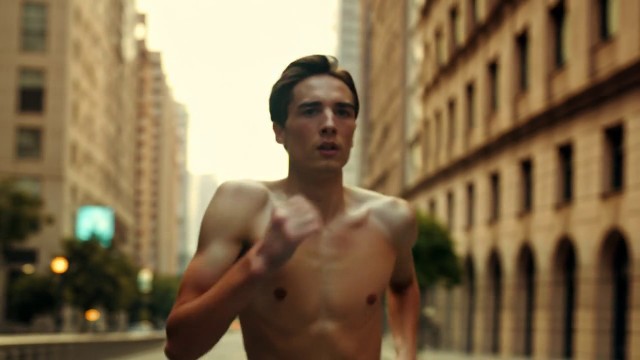} &
        \includegraphics[width=0.168\textwidth]{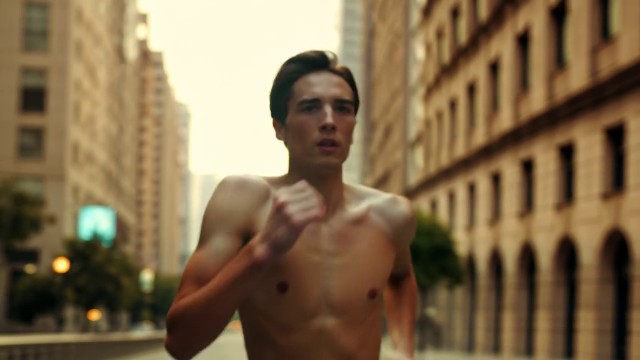} \\
    \end{tabular}
    \captionof{figure}{\textbf{Inter-frame flickering at an AR block boundary.} Across the block edge at frame $129$, Dummy Forcing's eviction pops in a spurious structure (a tree on the right, red box, frame $130$), while \textit{Full Attention} and \textit{HeadCast} stay temporally consistent.}
    \label{fig:flicker}
\end{figure*}

\subsection{Heterogeneous Cache Management}

Once the heads are classified, the monolithic KV cache is decoupled along the head dimension into three specialized, independent caches. With the $H$ heads partitioned into $N_{\text{sink}}$, $N_{\text{dummy}}$, $N_{\text{spatial}}$, and $N_{\text{global}}$ subsets, HeadCast restructures the memory footprint as follows:

\begin{itemize}
    \item \textbf{Sink-Dummy Cache ($\mathbf{K}_{\text{sd}}, \mathbf{V}_{\text{sd}}$):} Merges Sink and Dummy heads to halt temporal growth: the $N_{\text{sink}}$ heads freeze the initial block, while the $N_{\text{dummy}}$ heads keep only the most recent block via rolling overwrite.
    \item \textbf{Global Cache ($\mathbf{K}_{\text{global}}, \mathbf{V}_{\text{global}}$):} Maintains a standard sliding-window layout for the $N_{\text{global}}$ heads, retaining the initial sink tokens while evicting intermediate frames outside the window.
    \item \textbf{Spatial Cache ($\mathbf{K}_{\text{spatial}}, \mathbf{V}_{\text{spatial}}$):} Uses the same temporal layout as the Global Cache for the $N_{\text{spatial}}$ heads, but a localized spatial mask in the next phase sharply reduces its effective compute.
\end{itemize}

Slicing the monolithic tensor into these heterogeneous caches minimizes memory redundancy and yields contiguous layouts for downstream parallel computation.

\subsection{Head-Specific Attention}

At each autoregressive step, a lightweight kernel slices the queries along the head dimension and routes each group to its specialized path:

\begin{itemize}
    \item \textbf{Sink and Dummy Paths:} Queries fetch from the Sink-Dummy Cache. Because both types attend to exactly one historical block---the initial anchor for Sink, the latest block for Dummy---their KV sequences share an identical shape. We exploit this shared shape to fuse the two head types into a single attention kernel, eliminating their redundant long-range token matching.
    \item \textbf{Global Path:} Queries fetch from the Global Cache and perform standard sliding-window attention, bounding the temporal context length to $W$.
    \item \textbf{Spatial Path:} Queries access the full sliding window in time but under a block-to-block spatial constraint: the frame is partitioned into non-overlapping grids, and a query attends only to KV tokens in the same grid across all frames in the window (Figure~\ref{fig:spatial_head}). This decomposes the global attention matrix into many small, independent sub-matrices, which we stack along the batch dimension and compute with the standard FlashAttention kernel. The deployed grid intentionally differs from the sliding $(2r{+}1)\times(2r{+}1)$ neighborhood of the Spatial \emph{proxy}: a per-query sliding window cannot use FlashAttention and would need a slow custom FlexAttention kernel, whereas the non-overlapping grid yields independent dense sub-matrices that batch onto FlashAttention---a hardware-efficient approximation of the same locality.
\end{itemize}

Finally, the pathway outputs are concatenated back along the head dimension. By letting each head follow its own specialized path, HeadCast preserves visual quality while substantially accelerating inference.

\section{Experiments}

\subsection{Experimental Setup}

\textbf{Models and Baselines.} We plug HeadCast into four recent autoregressive video diffusion models: Self-Forcing~\cite{huang2025selfforcing}, LongLive~\cite{yang2026longlive}, Causal Forcing~\cite{zhu2026causalforcing}, and Reward Forcing~\cite{lu2026rewardforcing}. We compare against each model's full sliding-window attention and against Dummy Forcing, the most directly comparable training-free acceleration baseline.

\textbf{Evaluation Metrics.} For overall quality we report VBench~\cite{huang2024vbench} on 5-second clips (standard prompts) and VBench-Long on 30-second clips (the $128$ MovieGen~\cite{polyak2024moviegen} prompts of Self-Forcing++~\cite{cui2025selfforcingpp}), all refined with Qwen2.5-7B-Instruct~\cite{yang2024qwen25} following Self-Forcing~\cite{huang2025selfforcing}; every VBench score is averaged over $5$ seeds per prompt. As text-to-video generation has no ground-truth frames, we use each model's own full-attention output---the standard reference for an accelerated model---and report PSNR/LPIPS~\cite{zhang2018lpips} against it as a \emph{fidelity} axis complementing VBench's \emph{absolute}-quality axis. After aggregation VBench is relatively insensitive to the flickering and structural drift of aggressive eviction, which the fidelity axis and our qualitative comparisons expose. All measurements, including FPS and speedup, use a single GPU.

\textbf{Implementation Details.} For a fair comparison, the full-attention baseline and HeadCast share the same sliding window, with each block containing $3$ frames: a $4$-block window ($1$ fixed sink block plus $3$ rolling blocks) for $5$-second clips, and a $7$-block window ($1$ fixed sink block plus $6$ rolling blocks) for $30$-second clips. Online classification fires once the window first fills (e.g., autoregressive step $S_{\text{start}}=7$ in the $30$-second setting), at the maximum-noise denoising step $t=1000$. The cosine thresholds are tied to a single value $\theta_s=\theta_d=\theta_{\text{sp}}=0.95$, the spatial score threshold is $\theta_{\text{sc}}=0.755$, and the MSE penalty in the spatial score is weighted by $\gamma=5$. At classification time the Spatial proxy uses a $(2r{+}1)\times(2r{+}1)$ neighborhood with $r=2$, while the deployed Spatial path partitions each latent frame into non-overlapping $10\times 10$ cells (the boundary remainder is rounded). All reported speedups measure the post-classification, steady-state phase; the one-time classification is a separate fixed cost (under $1.5\%$ of total time on 30-second clips, $\sim$$5$--$8\%$ on 5-second clips; Appendix~\ref{sec:overhead}), so including it barely changes them.

\subsection{Main Results}

Table~\ref{tab:main_results} compares HeadCast, Dummy Forcing, and the full-attention baseline across the four backbones. On VBench Total, HeadCast stays within $0.15$ points of the full-attention baseline on every backbone, whereas Dummy Forcing loses up to $0.7$ points. Dynamic Degree, a coarse motion indicator, varies with backbone for both methods: HeadCast matches the baseline on Self-Forcing ($64.72$), stays close on Reward Forcing, is lower on LongLive ($29.84$ vs.\ $35.59$), and on Causal Forcing both accelerated methods read \emph{higher} than the baseline ($95.2$/$94.6$ vs.\ $85.0$). The two methods diverge far more sharply on frame-level fidelity, which VBench overlooks: against each model's full-attention output, HeadCast reaches $22$--$26$\,dB PSNR on the three $5$-second backbones against Dummy Forcing's $16$--$19$\,dB, and cuts LPIPS by $3$--$4\times$ ($0.045$--$0.068$ vs.\ $0.15$--$0.22$). This gap is exactly the inter-frame flickering Dummy Forcing's eviction introduces---structures popping in and out at block boundaries (Figure~\ref{fig:flicker})---which HeadCast avoids by retaining the Global heads. A block-boundary discontinuity metric in Appendix~\ref{sec:flicker_appendix} confirms this directly: on long clips Dummy Forcing raises the frame-to-frame jump at block edges by up to $36\%$ over full attention, whereas HeadCast matches the baseline.

These fidelity gains cost essentially nothing in speed. At standard resolution (480P) the KV cache is small, so attention is not yet dominant and speedups are modest for both methods (HeadCast $1.02$--$1.08\times$, Dummy Forcing $1.01$--$1.16\times$); only on LongLive ($30$\,s) does Dummy gain a meaningful FPS lead, yet it still sacrifices $0.75$\,dB PSNR. The real gains come at high resolution (\S\ref{sec:scalability}), where the KV cache grows and HeadCast's savings scale with it.

\subsection{Scalability to High Resolution}
\label{sec:scalability}

HeadCast's savings grow with resolution because the Spatial path uses a fixed-size grid: as each frame produces more tokens, the fraction of historical KV those heads access shrinks (Table~\ref{tab:high_res}). On Self-Forcing (5-second clips) the speedup rises from near $1\times$ at standard resolution to $1.31\times$ at 720P and $1.46\times$ at 1080P; on LongLive, whose 30-second horizon further enlarges the KV cache, it reaches $1.62\times$ and $1.95\times$.

Dummy Forcing scales more aggressively in raw FPS but its eviction degrades fidelity at both resolutions and backbones: on LongLive at 720P it trails HeadCast by $\sim$$2.4$\,dB PSNR ($12.83$ vs.\ $15.22$) and higher LPIPS ($0.455$ vs.\ $0.336$), with a larger gap on Self-Forcing ($18.81$ vs.\ $25.87$\,dB)---the conspicuous flickering of Figure~\ref{fig:flicker}, which HeadCast avoids while still delivering its $1.62$--$1.95\times$ speedup. HeadCast also shrinks the steady-state KV cache by $\sim$$33\%$ on Self-Forcing (Appendix~\ref{sec:memory}).

PSNR/LPIPS, measured against the full-attention output, are lower on long-video LongLive than on the $5$-second backbones ($15.22$ vs.\ $25.87$\,dB at 720P)---a property of the reference-based metric, not a loss of quality: over a $30$-second rollout the sparsified model diverges more from the full-attention trajectory, often making its own equally or more semantically faithful choices, while \emph{absolute} quality is preserved (VBench Total $79.05$ vs.\ $79.07$ at 720P, $77.26$ vs.\ $76.84$ at 1080P).

\begin{table}[t]
    \centering
    \footnotesize
    \setlength{\tabcolsep}{1.5pt}
    \begin{tabular}{@{} l c c c c c c @{}}
        \toprule
        \textbf{Method} & \textbf{Res.} & \textbf{FPS$\uparrow$} & \textbf{Spd.$\uparrow$} & \textbf{Total$\uparrow$} & \textbf{PSNR$\uparrow$} & \textbf{LPIPS$\downarrow$} \\
        \midrule
        Self-Forcing (5s) & 720P & 7.04 & 1.00$\times$ & 83.78 & $\infty$ & --- \\
        + Dummy Forcing & 720P & 9.60 & 1.36$\times$ & \textbf{83.91} & 18.81 & 0.2021 \\
        \textbf{+ HeadCast} & 720P & 9.21 & 1.31$\times$ & 83.83 & \textbf{25.87} & \textbf{0.0568} \\
        \hdashline \addlinespace[0.5ex]
        LongLive (30s) & 720P & 4.83 & 1.00$\times$ & 79.07 & $\infty$ & --- \\
        + Dummy Forcing & 720P & 8.40 & 1.74$\times$ & 78.85 & 12.83 & 0.4554 \\
        \textbf{+ HeadCast} & 720P & 7.82 & 1.62$\times$ & \textbf{79.05} & \textbf{15.22} & \textbf{0.3357} \\
        \midrule
        Self-Forcing (5s) & 1080P & 1.64 & 1.00$\times$ & 82.80 & $\infty$ & --- \\
        + Dummy Forcing & 1080P & 2.68 & 1.63$\times$ & 82.66 & 20.24 & 0.2059 \\
        \textbf{+ HeadCast} & 1080P & 2.40 & 1.46$\times$ & \textbf{82.90} & \textbf{27.17} & \textbf{0.0693} \\
        \hdashline \addlinespace[0.5ex]
        LongLive (30s) & 1080P & 1.07 & 1.00$\times$ & 76.84 & $\infty$ & --- \\
        + Dummy Forcing & 1080P & 2.31 & 2.16$\times$ & 77.18 & 12.29 & 0.5394 \\
        \textbf{+ HeadCast} & 1080P & 2.09 & 1.95$\times$ & \textbf{77.26} & \textbf{14.35} & \textbf{0.4198} \\
        \bottomrule
    \end{tabular}
    \caption{Scalability to 720P and 1080P on Self-Forcing (5-second clips) and LongLive (30-second clips). PSNR and LPIPS are computed against each model's full-attention output (hence ``$\infty$/---'' for the baseline).}
    \label{tab:high_res}
\end{table}

\subsection{User Study}
\label{sec:user_study}

Since aggregate VBench is insensitive to eviction-induced flickering (\S\ref{sec:scalability}), we additionally run a blind Two-Alternative Forced Choice (2AFC) human study (full protocol in Appendix~\ref{sec:user_study_protocol}): given two same-prompt videos, each participant makes a single \emph{overall-preference} choice, in two settings---HeadCast against the full-attention baseline and against Dummy Forcing. Table~\ref{tab:user_study} reports the share of comparisons favoring each method.

\begin{table}[t]
    \centering
    \footnotesize
    \setlength{\tabcolsep}{6pt}
    \begin{tabular}{@{}l c c@{}}
        \toprule
        \textbf{Comparison} & \textbf{HeadCast} & \textbf{Opponent} \\
        \midrule
        vs.\ Full Attention & \textbf{39.6\%} & 60.4\% \\
        vs.\ Dummy Forcing & \textbf{90.9\%} & 9.1\% \\
        \bottomrule
    \end{tabular}
    \caption{User study. Percentage of $2$AFC comparisons preferring each method (each row sums to $100\%$).}
    \label{tab:user_study}
\end{table}

The result is decisive against Dummy Forcing: HeadCast is preferred in $90.9\%$ of comparisons, reflecting the flicker-free quality that VBench alone fails to capture. Against the far stronger full-attention baseline, HeadCast is still preferred in $39.6\%$ of comparisons despite running substantially faster, indicating that its acceleration costs only a modest, often imperceptible quality margin.

\subsection{Ablation Studies}

Unless stated otherwise, ablations use Self-Forcing on $5$-second clips at 720P. We study four design choices: the contribution of each head archetype, the two classification thresholds, the denoising timestep $t$, and the stability of the one-shot classification.

\textbf{Effectiveness of Distinct Head Archetypes.} We ablate the two sparsity sources, grouping Sink and Dummy heads into \textit{Temporal Sparsity} and Spatial heads into \textit{Spatial Sparsity}, and activating them on top of the All-Global baseline (Table~\ref{tab:ablation_modules}).

\begin{table}[t]
    \centering
    \footnotesize
    \setlength{\tabcolsep}{2pt}
    \begin{tabular}{@{}l c c c c c@{}}
        \toprule
        \textbf{Configuration} & \textbf{Temp.} & \textbf{Spat.} & \textbf{Spd.$\uparrow$} & \textbf{PSNR$\uparrow$} & \textbf{VBench$\uparrow$} \\
        \midrule
        All Global (Baseline) & $\times$ & $\times$ & 1.00$\times$ & --- & 83.78 \\
        + Spatial only & $\times$ & \checkmark & 1.19$\times$ & 25.04 & 83.76 \\
        + Sink+Dummy only & \checkmark & $\times$ & 1.04$\times$ & 27.99 & 83.82 \\
        \midrule
        \textbf{Full HeadCast} & \checkmark & \checkmark & \textbf{1.31$\times$} & 25.87 & \textbf{83.83} \\
        \bottomrule
    \end{tabular}
    \caption{Ablation on head archetypes. Temp.\ groups Sink and Dummy heads; Spat.\ denotes Spatial heads.}
    \label{tab:ablation_modules}
\end{table}

Spatial Sparsity is the dominant speed lever ($1.19\times$) by fragmenting the dense attention map into independent local sub-matrices, but it bears most of the fidelity cost ($25.04$\,dB PSNR); Temporal Sparsity (Sink+Dummy) is gentler on both axes ($1.04\times$, $27.99$\,dB) by merely capping the attended sequence length. Combining them reaches $1.31\times$ with no measurable quality loss---VBench Total stays within $0.05$ points of the All-Global baseline.

\textbf{Impact of Classification Thresholds.} HeadCast is controlled by two thresholds: the cosine threshold $\theta\!:=\!\theta_s\!=\!\theta_d\!=\!\theta_{\text{sp}}$ that gates Sink/Dummy/Spatial assignment, and the spatial score threshold $\theta_{\text{sc}}$. We sweep each independently, measuring FPS and PSNR.

\begin{figure}[htbp]
    \centering
    \includegraphics[width=\columnwidth]{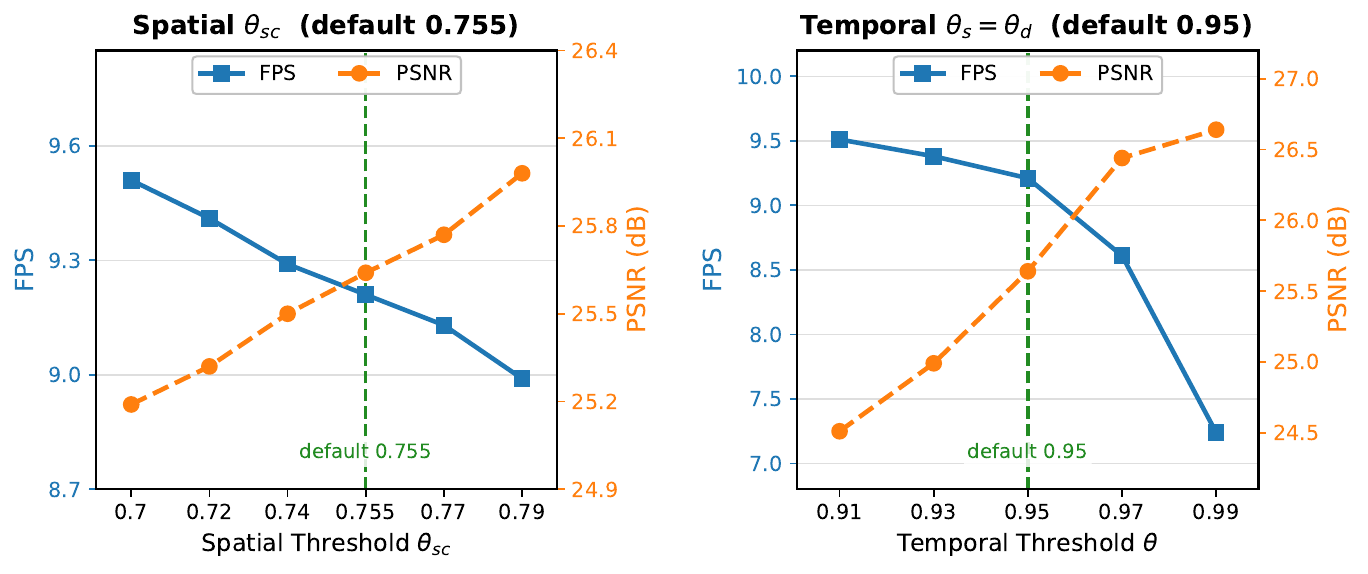}
    \caption{Ablation of classification thresholds. \textit{Left:} spatial score threshold $\theta_{\text{sc}}$. \textit{Right:} cosine threshold $\theta$. Lowering either threshold routes more heads to sparse paths, trading PSNR for FPS; green dashed lines mark our defaults ($\theta_{\text{sc}}{=}0.755$, $\theta{=}0.95$).}
    \label{fig:ablation_thresholds}
\end{figure}

As shown in Figure~\ref{fig:ablation_thresholds}, tightening $\theta$ from $0.91$ to $0.99$ raises PSNR from $24.51$ to $26.64$\,dB at the price of dropping FPS from $9.51$ to $7.24$, while $\theta_{\text{sc}}$ controls a milder trade-off ($\pm 0.52$ FPS, $\mp 0.79$\,dB). Both curves are smooth and monotone, and our defaults strike a favorable balance between fidelity and speed. Because the thresholds compare \emph{cosine similarities} of attention patterns---quantities that are normalized and largely architecture-agnostic---we apply this single set ($\theta{=}0.95$, $\theta_{\text{sc}}{=}0.755$) unchanged across all four backbones and all resolutions, with no per-model retuning.

\textbf{Impact of the Classification Timestep.} We examine the denoising timestep $t$ at which the one-shot classification is performed (Table~\ref{tab:ablation_timestep}).

\begin{table}[htbp]
    \centering
    \footnotesize
    \setlength{\tabcolsep}{2.5pt}
    \begin{tabular}{@{}c c c c c c c@{}}
        \toprule
        \textbf{$t$} & \textbf{Sink} & \textbf{Dummy} & \textbf{Spatial} & \textbf{Global} & \textbf{PSNR$\uparrow$} & \textbf{LPIPS$\downarrow$} \\
        \midrule
        \textbf{1000} & 29.4 & \textbf{136.3} & 84.1 & \textbf{110.2} & \textbf{25.87} & \textbf{0.0568} \\
        750 & 41.3 & 149.4 & 81.7 & 87.7 & 25.19 & 0.0612 \\
        500 & 48.1 & 166.3 & 69.3 & 76.2 & 24.79 & 0.0649 \\
        250 & 53.0 & 190.8 & 50.4 & 65.9 & 24.52 & 0.0674 \\
        \bottomrule
    \end{tabular}
    \caption{Ablation on the classification denoising timestep $t$: lower $t$ routes more heads to \textit{Dummy} and fewer to \textit{Global}, so fidelity degrades and $t=1000$ is best. Head counts are out of $360$ (means; may not sum to exactly $360$ due to rounding).}
    \label{tab:ablation_timestep}
\end{table}

As denoising proceeds from $t=1000$ toward lower noise levels, the model concentrates on local, high-frequency detail, so progressively more heads attend only to their own current frame---the defining behavior of the \textit{Dummy} archetype, which our decision hierarchy assigns with priority. The Dummy count therefore grows steadily as $t$ falls, with a smaller rise in Sink heads, while the number of \textit{Global} heads---those carrying the long-range temporal context---collapses. Heads that genuinely require global history are thereby misrouted to the constant-size Dummy/Sink window and lose the keys needed for temporal consistency, introducing inter-frame flickering; both PSNR and LPIPS degrade monotonically as $t$ drops. Classifying at the maximum noise level ($t=1000$), where the most heads still expose their full long-range structure, therefore yields the best fidelity and supports our default.

\begin{figure}[htbp]
    \centering
    \includegraphics[width=0.85\columnwidth]{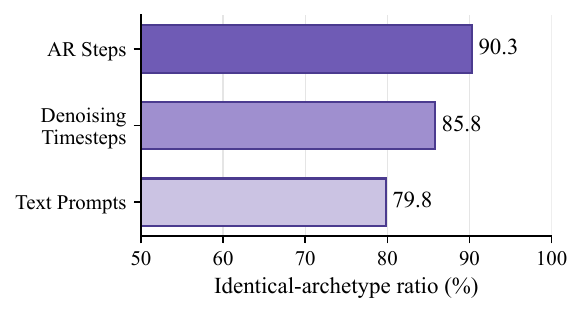}
    \caption{Stability of the one-shot classification. Each bar is the fraction of the $360$ heads that keep the \emph{identical} four-way archetype when the classification is repeated across that axis (mean over all sampled pairs).}
    \label{fig:stability_bar}
\end{figure}

\looseness=-1
\textbf{Stability of the One-Shot Classification.} HeadCast classifies each head exactly once and reuses that assignment for the entire rollout, so we verify that the assignment is robust to the conditions under which it is computed: the autoregressive (AR) step, the denoising timestep, and the text prompt. For each axis we re-run the classification along it and measure the fraction of the $360$ heads that receive the \emph{identical} four-way archetype (Figure~\ref{fig:stability_bar}). The assignment is highly consistent on every axis---$90.3\%$ across AR steps (frame $12$ vs.\ $18$, two blocks apart), $85.8\%$ across denoising timesteps, and $79.8\%$ across text prompts---confirming Observation~3: a single classification at a representative AR step and the maximum-noise timestep suffices, and re-profiling at every step is unnecessary.

\section{Conclusion}

\looseness=-1
We presented HeadCast, a training-free framework that classifies each attention head once---at the maximum-noise step---into one of four archetypes (Sink, Dummy, Spatial, Global) and routes it to a head-specific path. Because the Spatial path uses a fixed-size grid, its savings grow with resolution, reaching $1.62\times$ at 720P and $1.95\times$ at 1080P, while the retained Global heads preserve the temporal consistency that aggressive eviction destroys. Across state-of-the-art AR models it matches full-attention VBench quality at markedly higher frame-level fidelity than prior training-free eviction---making head-level heterogeneity a practical lever for efficient AR video generation.

\bibliography{mybibliography}

\clearpage
\appendix

\noindent{\LARGE\bfseries Appendix}\par
\vspace{8pt}

\noindent In this appendix, we provide additional details and results, including:
\begin{itemize}
    \item Per-head attention-pattern visualizations (Appendix~\ref{sec:perhead_vis}).
    \item The full head-archetype distribution and a theoretical FLOPs analysis (Appendix~\ref{sec:archetype_dist}).
    \item The KV-cache memory cost of each method (Appendix~\ref{sec:memory}).
    \item The one-time classification overhead (Appendix~\ref{sec:overhead}).
    \item A block-boundary discontinuity metric and further inter-frame flicker comparisons (Appendix~\ref{sec:flicker_appendix}).
    \item Additional qualitative results on Self-Forcing ($5$\,s) and LongLive ($30$\,s) (Appendix~\ref{sec:qualitative}).
    \item The user study protocol (Appendix~\ref{sec:user_study_protocol}).
\end{itemize}

\section{Per-Head Attention Pattern Visualization}
\label{sec:perhead_vis}

\begin{figure*}[p]
  \centering
  \setlength{\tabcolsep}{1pt}
  \begin{tabular}{ccc}
    \multicolumn{3}{c}{\textbf{Sink head}} \\[2pt]
    \headrows{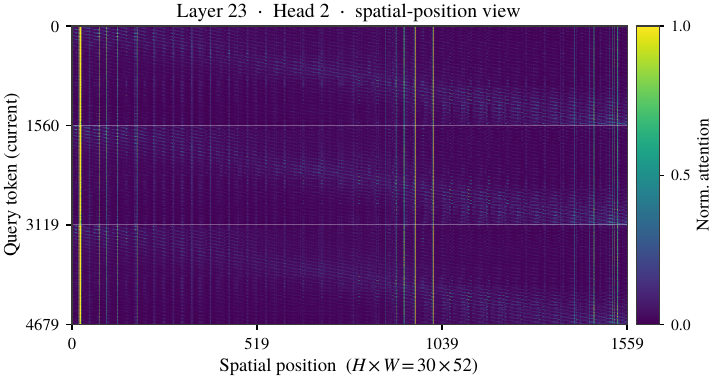}
    \multicolumn{3}{c}{\textbf{Dummy head}} \\[2pt]
    \headrows{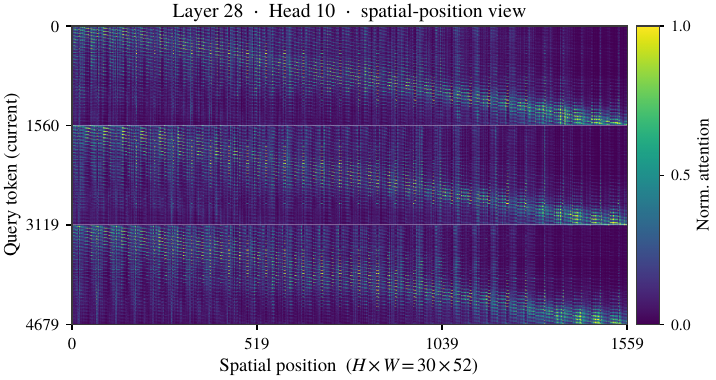}
    \multicolumn{3}{c}{\textbf{Spatial head}} \\[2pt]
    \headrows{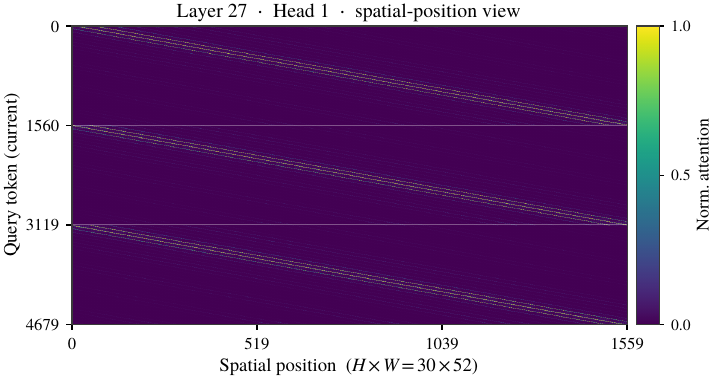}
  \end{tabular}
  \caption{\textbf{Sink, Dummy, and Spatial heads.} One representative head per archetype, each occupying two rows (frame~6 then frame~18). Columns: raw map, per-frame (temporal) aggregation, per-position (spatial) aggregation. The \emph{Sink} head (Layer~23, Head~2) concentrates on the first cached block; the \emph{Dummy} head (Layer~28, Head~10) collapses onto the current block; the \emph{Spatial} head (Layer~27, Head~1) attends to a fixed local neighborhood.}
  \label{fig:gallery_sds}
\end{figure*}

\begin{figure*}[t]
  \centering
  \setlength{\tabcolsep}{1pt}
  \begin{tabular}{ccc}
    \multicolumn{3}{c}{\textbf{Global head}} \\[2pt]
    \headrows{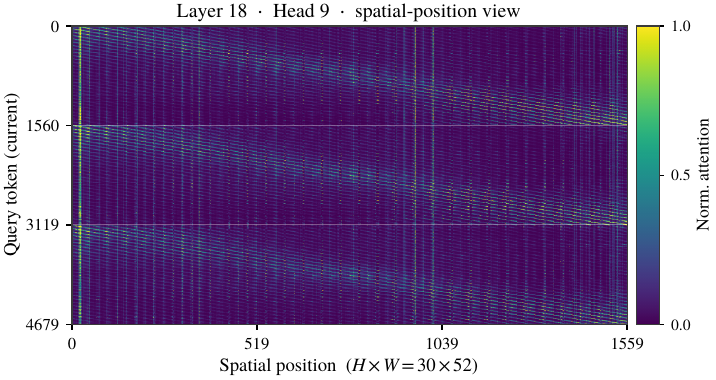}
  \end{tabular}
  \caption{\textbf{Global head.} One representative head occupying two rows (frame~6 then frame~18). Columns: raw map, per-frame (temporal) aggregation, per-position (spatial) aggregation. The \emph{Global} head (Layer~18, Head~9) spreads its attention broadly across the whole history.}
  \label{fig:gallery_global}
\end{figure*}

To complement the four head archetypes of Figure~\ref{fig:horizontal_four}, we enlarge one representative head per archetype---each different from the examples there---in Figures~\ref{fig:gallery_sds} and~\ref{fig:gallery_global}. Each head occupies two rows (top: decoding frame~$6$; bottom: frame~$18$), with three columns: the \textit{raw} attention map, its per-frame (\textit{temporal}) aggregation, and its per-position (\textit{spatial}) aggregation, all at the maximum-noise step ($t=1000$). Comparing the two rows shows that each head keeps its archetype from frame~$6$ to frame~$18$---a visual counterpart to the quantitative stability in Figure~\ref{fig:stability_bar} that licenses our \emph{one-shot} classification at a single denoising step.

\noindent\textbf{Sink heads} (Figure~\ref{fig:gallery_sds}, top) concentrate a disproportionate share of their attention mass on the \emph{first} cached block, while the spatial view exhibits no coherent locality structure---the mass falls on a few scattered spatial positions---so these heads contribute little position-specific information to the current step. For the shown head (Layer~23, Head~2) the leading frame of this block alone receives $38\%$ of the mass at decoding frame~$6$ and remains by far the dominant frame at decoding frame~$18$ ($15\%$, roughly $3\times$ the uniform share).

\noindent\textbf{Dummy heads} (Figure~\ref{fig:gallery_sds}, middle) collapse onto the \emph{current} block: essentially all attention mass falls on its frames ($\sim$$33\%$ each), with each query attending almost exclusively to its own frame and negligible weight on older history, so these heads read only a constant-size tail of the cache and are among the cheapest to serve.

\noindent\textbf{Spatial heads} (Figure~\ref{fig:gallery_sds}, bottom) do not lock onto a single frame; instead, the raw map reveals a banded, near-diagonal structure in which each query token attends to a fixed local neighborhood replicated across the cached frames. The per-position (spatial) view makes this locality explicit.

\noindent\textbf{Global heads} (Figure~\ref{fig:gallery_global}) spread their attention broadly across all cached frames and all spatial positions, with no single frame exceeding roughly one-fifth of the total mass. These heads carry the long-range temporal context that HeadCast explicitly preserves by routing them to the full-attention path.

\section{Head Archetype Distribution and Theoretical Compute Savings}
\label{sec:archetype_dist}

We report the full four-way head distribution under the default configuration and use it to attribute HeadCast's speedup to its constituent archetypes. Unlike the merged \texttt{Temp.}/\texttt{Spat.}\ columns of Table~\ref{tab:ablation_modules} (which fold Sink and Dummy into a single \emph{Temporal} group), Table~\ref{tab:archetype_dist} gives the complete Sink/Dummy/Spatial/Global counts at the default $t{=}1000$ for LongLive across resolutions---the running example for the FLOPs analysis below; all $30\times12=360$ heads are accounted for.

\begin{table}[htbp]
    \centering
    \footnotesize
    \setlength{\tabcolsep}{5pt}
    \begin{tabular}{@{}l c c c c@{}}
        \toprule
        \textbf{Resolution} & \textbf{Sink} & \textbf{Dummy} & \textbf{Spatial} & \textbf{Global} \\
        \midrule
        480P  & 1.6 & 93.8 & 168.2 & 96.5 \\
        720P  & 1.9 & 98.2 & 167.2 & 92.7 \\
        1080P & 1.5 & 91.7 & 203.2 & 63.7 \\
        \bottomrule
    \end{tabular}
    \caption{Head archetype distribution (LongLive, default $t{=}1000$). Counts are out of $360$ heads ($30$ layers $\times$ $12$ heads); entries are means and may not sum to exactly $360$ due to rounding.}
    \label{tab:archetype_dist}
\end{table}

\textbf{Theoretical compute savings.} The attention FLOPs of a single head are dominated by its two matmuls---$QK^\top$ and $\text{softmax}\,V$---totalling $4\,B\,L_q\,L_k\,d_{\text{head}}$. Because every comparison below is a ratio, this constant factor cancels, so we drop it and write the cost as $B\,L_q\,L_k\,d_{\text{head}}$. Under our inference configuration each block has sequence length $L$, so a query block has $L_q=L$ and attends to a window of $L_k=WL$ tokens. Taking LongLive's $30$-second sliding window as a concrete example ($W{=}8$: $7$ cached blocks plus the current block), a full-attention (Global) head therefore costs $8BL^2d$. Each archetype reduces this differently:

\begin{itemize}
    \item \textbf{Sink} attends only to the first (sink) block and the current block: $8BL^2d \rightarrow 2BL^2d$ ($75\%$ reduction).
    \item \textbf{Dummy} attends only to the previous block and the current block: $8BL^2d \rightarrow 2BL^2d$ ($75\%$ reduction).
    \item \textbf{Spatial} restricts each query to its own fixed $10\times10$ cell, which at 480P (a $30\times52$ latent grid) covers $\approx\tfrac{1}{16}$ of the frame: $8BL^2d \rightarrow \tfrac12 BL^2d$ ($\approx$$94\%$ reduction at 480P, and more at higher resolution, where the fixed-size cell spans a smaller fraction).
    \item \textbf{Global} is unchanged: $8BL^2d$ ($0\%$ reduction).
\end{itemize}

Integrating over the measured distribution, the attention-FLOPs ratio is $\frac{2N_{\text{s}}+2N_{\text{d}}+\frac12 N_{\text{sp}}+8N_{\text{g}}}{8\,N_{\text{total}}}$, giving a \textbf{$\sim$$64\%$} reduction at 480P that grows at higher resolution, as the fixed-size Spatial cells cover a smaller fraction of each frame. This exceeds the measured per-block speedup ($1.62$--$1.95\times$, i.e.\ a $38$--$49\%$ wall-clock reduction), which is reported on post-classification blocks and therefore contains no classification cost. The gap to the theoretical FLOPs saving has two sources: attention is only part of the per-block compute (the diffusion MLP and projection layers are unaffected), and HeadCast dispatches its archetype groups through separate attention kernels---about three launches per block instead of one---so the added kernel-launch overhead offsets part of the FLOPs reduction.

\section{KV-Cache Memory Cost}
\label{sec:memory}

\begin{figure}[htbp]
    \centering
    \includegraphics[width=\columnwidth]{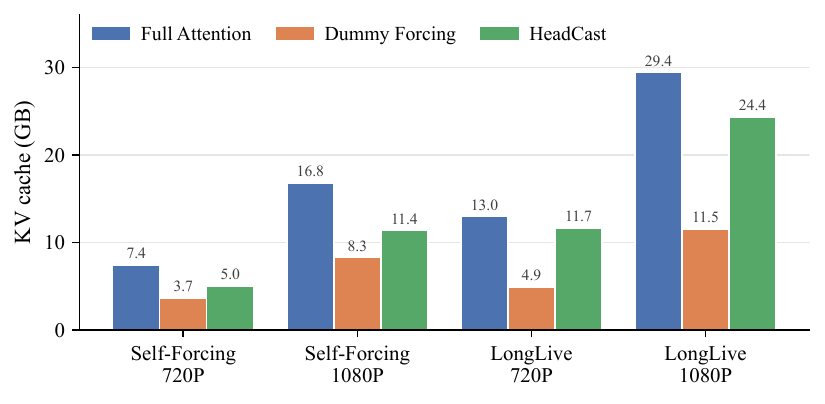}
    \caption{Steady-state KV-cache size (GB) for the full-attention baseline, Dummy Forcing, and HeadCast. HeadCast caches less than full attention at every setting (about $33\%$ less on Self-Forcing) while retaining substantially more history than Dummy Forcing.}
    \label{fig:kv_bar}
\end{figure}

A KV-cache compression method should also shrink memory, so we report the steady-state KV-cache size for the full-attention baseline, Dummy Forcing, and HeadCast (Figure~\ref{fig:kv_bar}). HeadCast consistently caches less than full attention---about $33\%$ less on Self-Forcing---by capping the history that Sink and Dummy heads retain, while still keeping substantially more context than Dummy Forcing's aggressive eviction, which is precisely what spares it the inter-frame flicker that eviction produces.

\textbf{Theoretical KV-cache reduction.} With a $W$-block window ($W{=}8$), a full head caches $WL$ tokens. HeadCast retains KV per archetype: Sink and Dummy heads keep only $2L$ tokens (their anchor block plus the current block, a $75\%$ reduction), while Spatial and Global heads still require the full window (Spatial partitions only the spatial dimensions, dropping no frames). The expected steady-state KV reduction is therefore $\frac{N_{\text{s}}+N_{\text{d}}}{N_{\text{total}}}\times0.75$. For the Self-Forcing distribution ($N_{\text{s}}{+}N_{\text{d}}\approx166/360\approx46\%$) this predicts a $\sim34\%$ reduction, close to the measured $\sim33\%$ in Figure~\ref{fig:kv_bar}. LongLive, whose Sink$+$Dummy share is smaller, shows a correspondingly milder reduction.

\section{Classification Overhead}
\label{sec:overhead}

HeadCast's only method-specific extra cost is the one-time classification, which fires on a single autoregressive block. Our reported speedups are measured on the post-classification (steady-state) blocks and therefore exclude this one-time cost; Table~\ref{tab:overhead} quantifies it separately, as a fraction of the total diffusion time. It is small and, crucially, \emph{amortizes with the rollout length}: on $5$-second Self-Forcing clips it is $\sim$$5$--$8\%$, but on the flagship $30$-second LongLive setting it drops below $1.5\%$---only $0.6\%$ at 1080P. Folding this one-time classification into an end-to-end figure would thus lower the speedup only marginally, and negligibly on long rollouts.

\begin{table}[htbp]
    \centering
    \footnotesize
    \setlength{\tabcolsep}{6pt}
    \begin{tabular}{@{}l c c c@{}}
        \toprule
        \textbf{Backbone (length)} & \textbf{480P} & \textbf{720P} & \textbf{1080P} \\
        \midrule
        Self-Forcing (5\,s)  & 7.9\% & 7.1\% & 4.8\% \\
        LongLive (30\,s)     & 1.3\% & 1.0\% & \textbf{0.6\%} \\
        \bottomrule
    \end{tabular}
    \caption{One-time classification overhead as a fraction of total diffusion time. The overhead amortizes as the video lengthens, becoming negligible on the $30$-second setting.}
    \label{tab:overhead}
\end{table}

\section{Inter-Frame Flickering}
\label{sec:flicker_appendix}

\textbf{A block-boundary discontinuity metric.} Eviction-induced flicker is \emph{localized}: it appears as an abrupt content change exactly at the autoregressive block boundaries, where the sliding cache is updated and stale keys are dropped. We measure it directly. For consecutive decoded frames we compute the mean absolute pixel difference $d_t=\frac{1}{HWC}\sum|f_t-f_{t-1}|$, and define the \emph{block-boundary discontinuity} (BBD) as the ratio of the average $d_t$ over block-boundary frames to the average $d_t$ within blocks. A temporally smooth clip has $\mathrm{BBD}\approx1$; values above $1$ mean the boundaries jump more than ordinary motion---the pop-in/pop-out artifact of Figure~\ref{fig:flicker}. Because the ratio is localized at boundaries, it isolates exactly the failure mode that frame-averaged scores such as VBench dilute away.

\begin{table}[htbp]
    \centering
    \footnotesize
    \setlength{\tabcolsep}{4pt}
    \begin{tabular}{@{}l c c c c@{}}
        \toprule
        \textbf{Setting} & \textbf{Full} & \textbf{HeadCast} & \textbf{Dummy} & \textbf{$\Delta_{\text{Dummy}}$} \\
        \midrule
        Self-Forcing (5s), 480P  & 1.192 & 1.192 & 1.323 & $+0.131$ \\
        Self-Forcing (5s), 720P  & 1.235 & 1.238 & 1.272 & $+0.037$ \\
        Self-Forcing (5s), 1080P & 1.265 & 1.250 & 1.285 & $+0.020$ \\
        \midrule
        LongLive (30s), 480P  & 1.304 & 1.260 & 1.777 & $+0.473$ \\
        LongLive (30s), 720P  & 1.245 & 1.242 & 1.536 & $+0.291$ \\
        LongLive (30s), 1080P & 1.232 & 1.166 & 1.395 & $+0.163$ \\
        \bottomrule
    \end{tabular}
    \caption{\textbf{Block-boundary discontinuity} (BBD; $1.0$ = no boundary jump, lower is smoother). HeadCast matches the full-attention baseline at every setting, whereas Dummy Forcing's eviction inflates the discontinuity---most severely on the $30$-second LongLive clips. $\Delta_{\text{Dummy}}$ is Dummy minus Full, computed at full precision (so it may differ from the difference of the rounded entries by $0.001$).}
    \label{tab:bbd}
\end{table}

Table~\ref{tab:bbd} isolates the effect of cache eviction by comparing each accelerated method against full attention. HeadCast adds essentially \emph{no} boundary discontinuity: its BBD equals the full-attention baseline on the $5$-second clips and is even slightly \emph{lower} on the $30$-second clips (HeadCast\,$-$\,Full\,$\le\!0$ throughout). Dummy Forcing instead inflates it, and the inflation scales with the rollout length---which is exactly what one expects of an artifact that accumulates as the cache is repeatedly evicted. On $5$-second clips, where eviction happens only a handful of times, the excess is small ($+0.02$ to $+0.13$); over the $30$-second horizon it compounds to $+0.16$--$+0.47$ ($13$--$36\%$ relative), peaking on the flagship long-video LongLive setting. Flicker is thus fundamentally a long-rollout problem, and it is precisely there that HeadCast's explicit Global heads pay off. This trend mirrors the qualitative examples below and the PSNR/LPIPS gap in Table~\ref{tab:main_results}---an artifact the metric and figures expose but that aggregate VBench dilutes away.

To complement the example in Figure~\ref{fig:flicker}, we provide two further side-by-side comparisons that isolate the inter-frame flickering induced by aggressive cache eviction. Each example spans a five-frame window centered on an autoregressive block boundary, where Dummy Forcing's coarse eviction is most prone to hallucinating or dropping structure. As before, all three methods (Dummy Forcing, Full Attention, and our HeadCast) are run from the same prompt on LongLive at 720P, and the frame index is shared across rows so that a given column shows the same timestamp under each method. The red box marks the region of interest on the Dummy Forcing row.

In both examples, Full Attention and HeadCast keep the highlighted structure temporally stable across the boundary, whereas Dummy Forcing introduces an abrupt, frame-to-frame change---precisely the artifact that lowers its PSNR/LPIPS yet, once VBench's temporal sub-dimensions are pooled into the aggregate score, is largely washed out.

\newcommand{\flrowlab}[1]{{\footnotesize\shortstack{#1}}}
\begin{figure*}[t]
    \centering
    \begin{subfigure}{\textwidth}
        \centering
        \setlength{\tabcolsep}{1.5pt}
        \renewcommand{\arraystretch}{0.5}
        \begin{tabular}{@{}c@{\hskip 2pt}ccccc@{}}
            & \footnotesize Frame 103 & \footnotesize Frame 104 & \footnotesize Frame 105 & \footnotesize Frame 106 & \footnotesize Frame 107 \\
            \flrowlab{Dummy\\Forcing} &
            \includegraphics[width=0.18\textwidth]{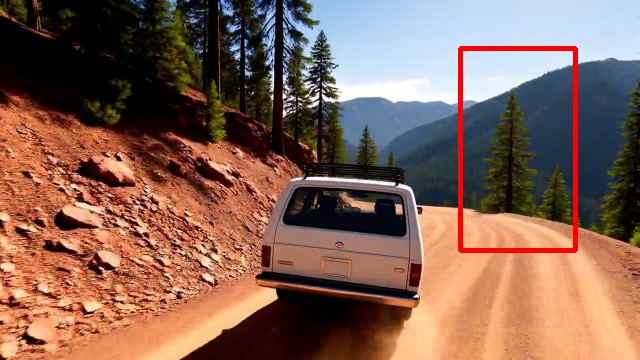} &
            \includegraphics[width=0.18\textwidth]{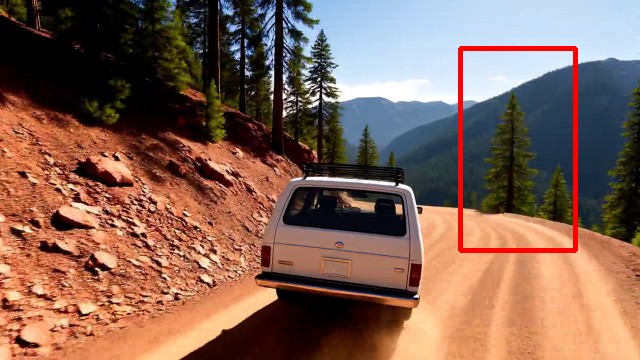} &
            \includegraphics[width=0.18\textwidth]{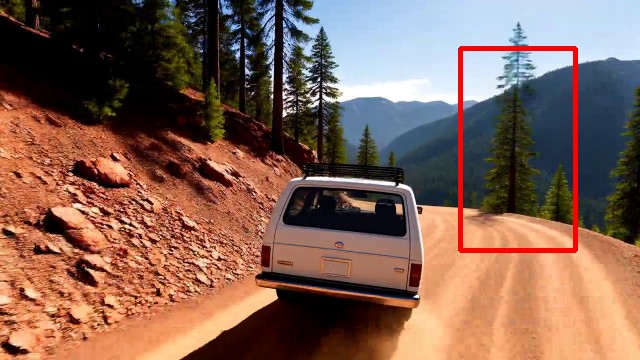} &
            \includegraphics[width=0.18\textwidth]{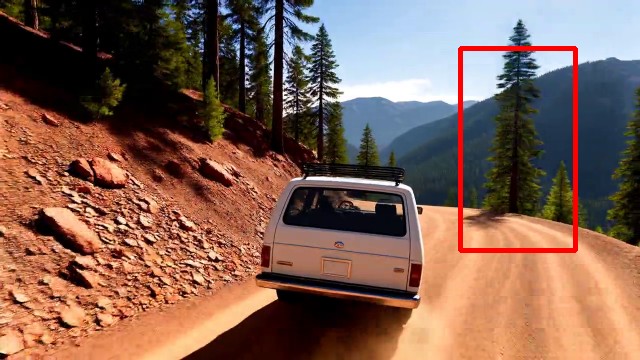} &
            \includegraphics[width=0.18\textwidth]{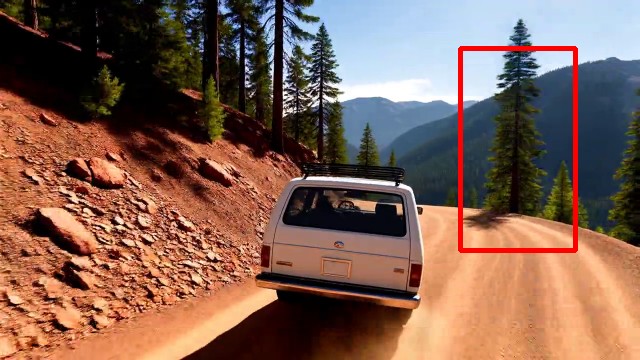} \\
            \flrowlab{Full\\Attention} &
            \includegraphics[width=0.18\textwidth]{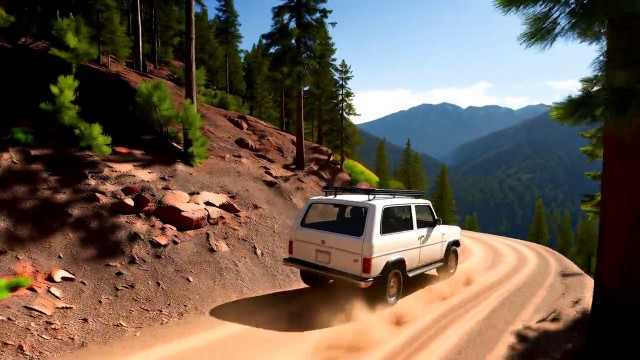} &
            \includegraphics[width=0.18\textwidth]{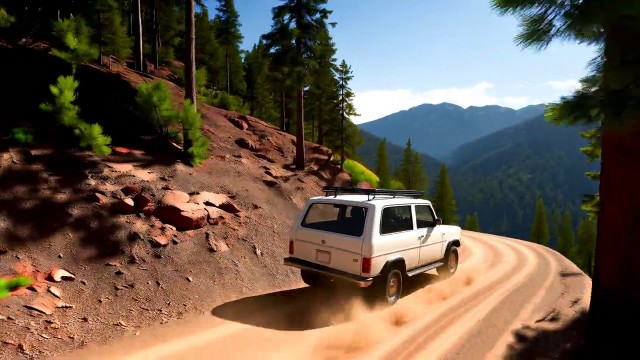} &
            \includegraphics[width=0.18\textwidth]{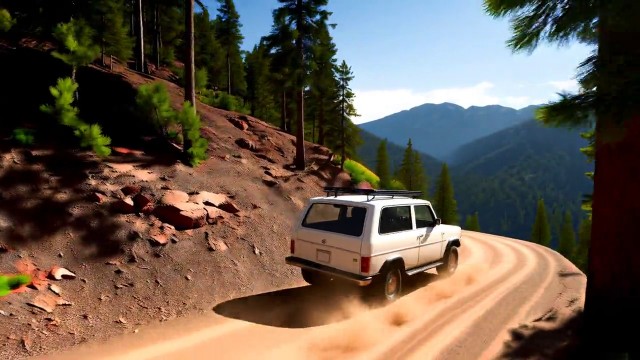} &
            \includegraphics[width=0.18\textwidth]{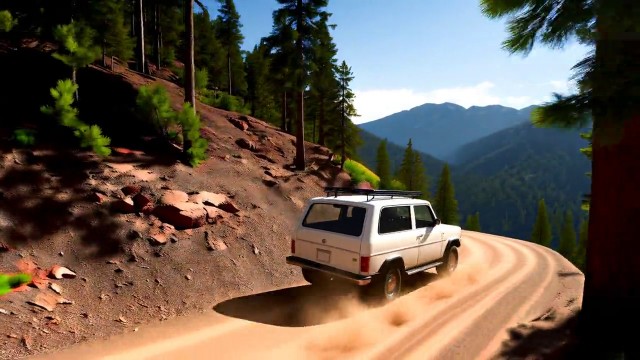} &
            \includegraphics[width=0.18\textwidth]{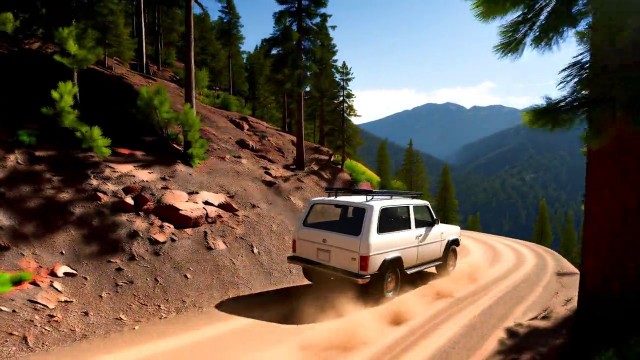} \\
            \flrowlab{HeadCast} &
            \includegraphics[width=0.18\textwidth]{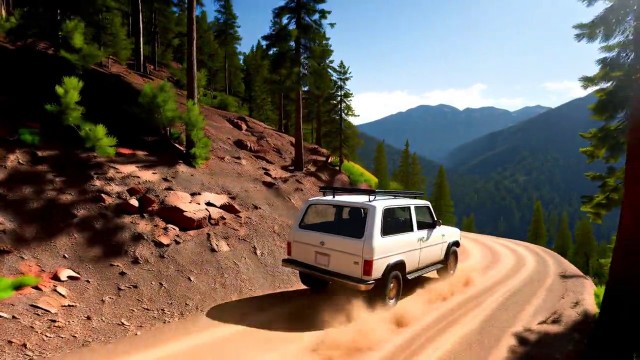} &
            \includegraphics[width=0.18\textwidth]{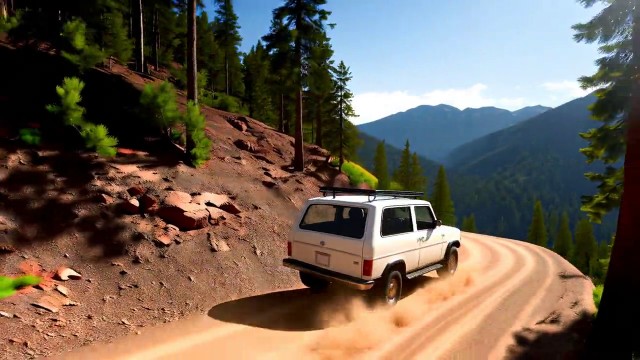} &
            \includegraphics[width=0.18\textwidth]{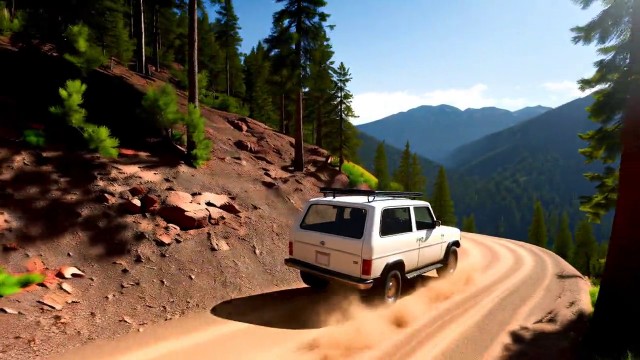} &
            \includegraphics[width=0.18\textwidth]{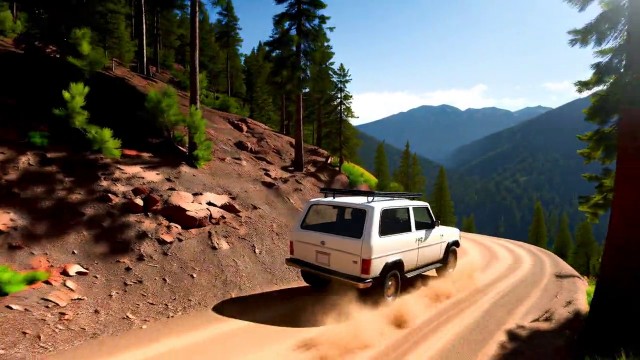} &
            \includegraphics[width=0.18\textwidth]{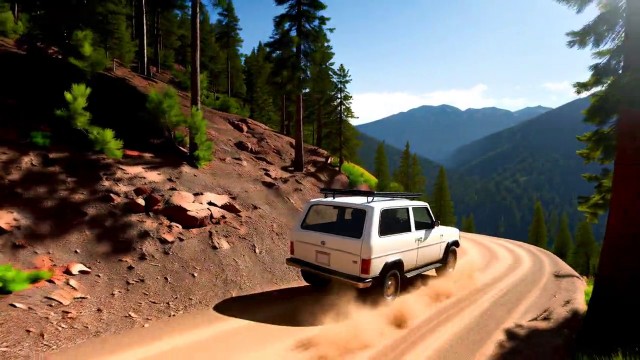} \\
        \end{tabular}
        \caption{Mountain-road scene: a pine tree on the right (red box) abruptly pops in and out across the block boundary at frame~$105$ under Dummy Forcing.}
        \label{fig:flicker_supp_tree}
    \end{subfigure}

    \vspace{8pt}

    \begin{subfigure}{\textwidth}
        \centering
        \setlength{\tabcolsep}{1.5pt}
        \renewcommand{\arraystretch}{0.5}
        \begin{tabular}{@{}c@{\hskip 2pt}ccccc@{}}
            & \footnotesize Frame 79 & \footnotesize Frame 80 & \footnotesize Frame 81 & \footnotesize Frame 82 & \footnotesize Frame 83 \\
            \flrowlab{Dummy\\Forcing} &
            \includegraphics[width=0.18\textwidth]{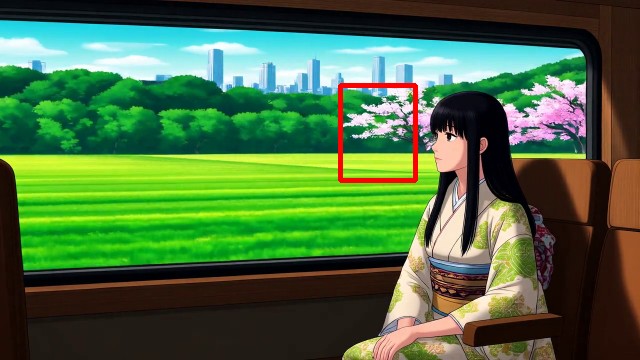} &
            \includegraphics[width=0.18\textwidth]{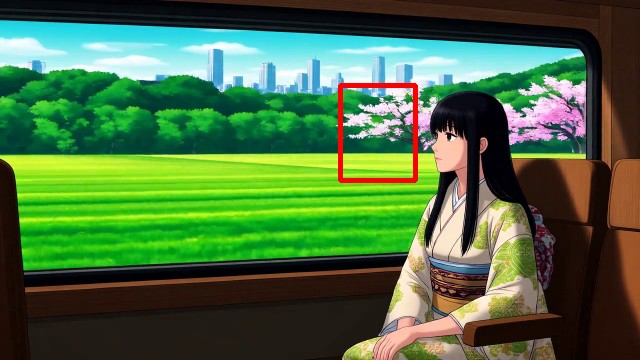} &
            \includegraphics[width=0.18\textwidth]{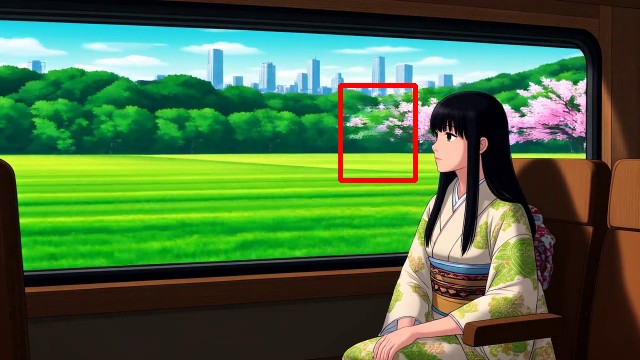} &
            \includegraphics[width=0.18\textwidth]{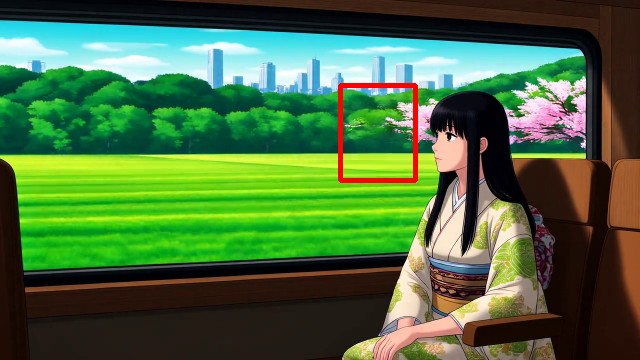} &
            \includegraphics[width=0.18\textwidth]{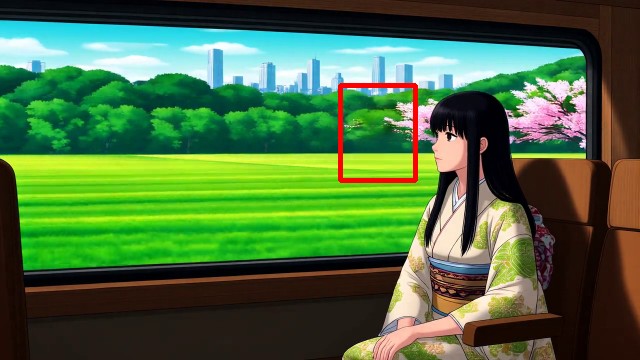} \\
            \flrowlab{Full\\Attention} &
            \includegraphics[width=0.18\textwidth]{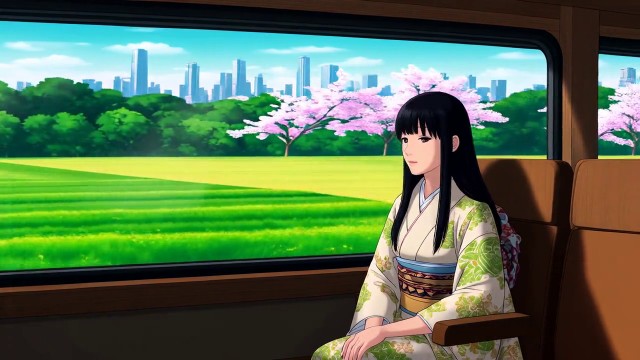} &
            \includegraphics[width=0.18\textwidth]{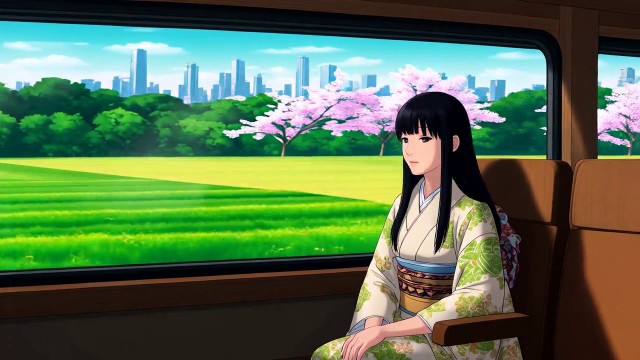} &
            \includegraphics[width=0.18\textwidth]{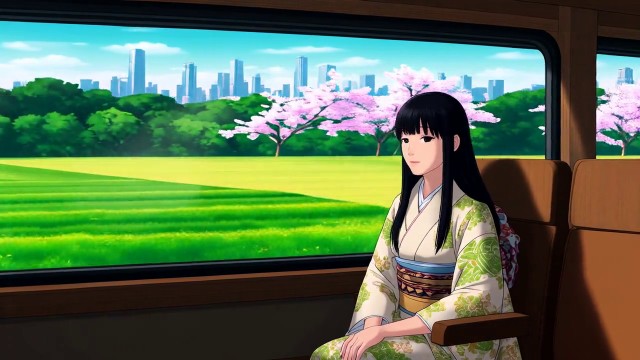} &
            \includegraphics[width=0.18\textwidth]{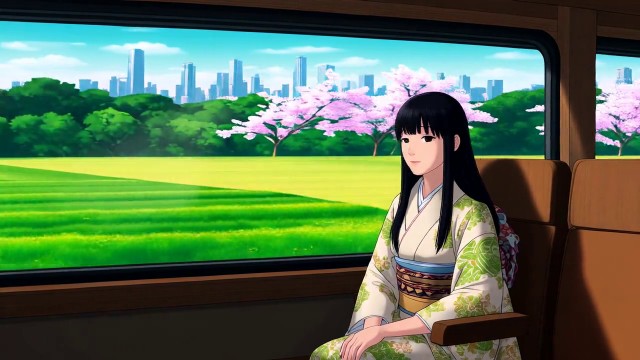} &
            \includegraphics[width=0.18\textwidth]{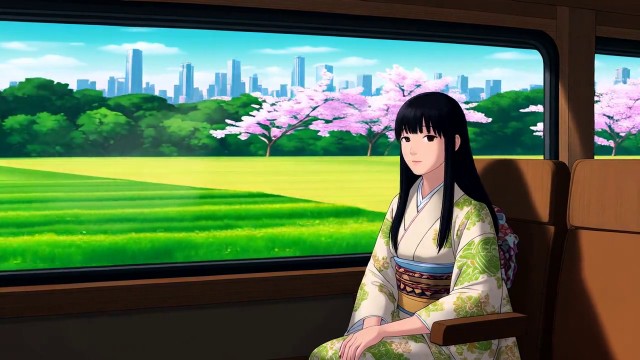} \\
            \flrowlab{HeadCast} &
            \includegraphics[width=0.18\textwidth]{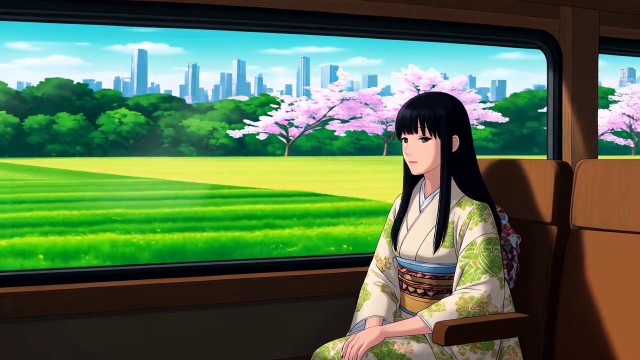} &
            \includegraphics[width=0.18\textwidth]{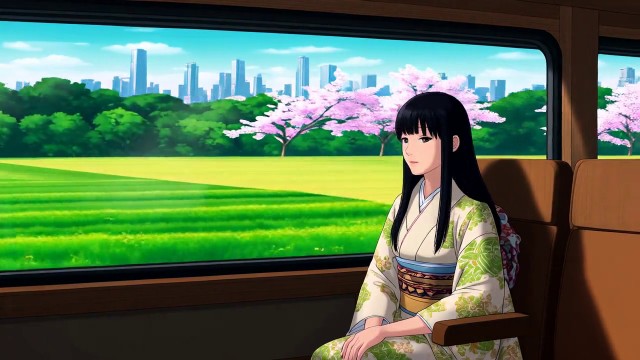} &
            \includegraphics[width=0.18\textwidth]{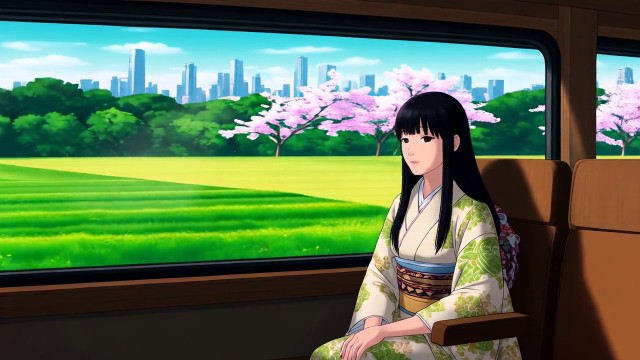} &
            \includegraphics[width=0.18\textwidth]{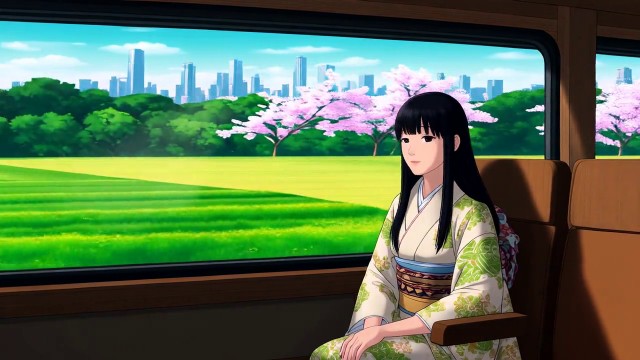} &
            \includegraphics[width=0.18\textwidth]{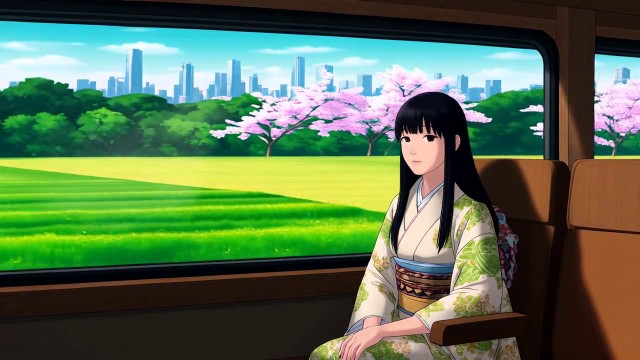} \\
        \end{tabular}
        \caption{Train-window scene: cherry blossoms seen through the window (red box) flicker across the block boundary at frame~$81$ under Dummy Forcing.}
        \label{fig:flicker_supp_blossom}
    \end{subfigure}
    \caption{\textbf{Additional inter-frame flickering examples at AR block boundaries} (LongLive, $30$\,s, 720P). Each five-frame window is centered on a block edge. Dummy Forcing's aggressive cache eviction introduces an abrupt, frame-to-frame change in the boxed region, whereas \textit{Full Attention} and our \textit{HeadCast} keep the scene temporally consistent.}
    \label{fig:flicker_supp}
\end{figure*}

\section{Additional Qualitative Results}
\label{sec:qualitative}

We compare HeadCast with the full-attention model frame by frame: in each block the top row is full attention and the bottom row is HeadCast, with five frames sampled at equal intervals. All prompts are drawn from the MovieGen~\cite{polyak2024moviegen} prompt set (the same suite used for VBench-Long evaluation in the main paper). On $5$-second Self-Forcing (720P; Figures~\ref{fig:results_supp_sf_a} and~\ref{fig:results_supp_sf_b})---a neon-lit Tokyo street, ships in a coffee cup, puppies in the snow---the two rows are nearly identical despite HeadCast's much smaller KV cache.

\newcommand{\resrowsS}[1]{%
    & \footnotesize Frame 0 & \footnotesize Frame 20 & \footnotesize Frame 40 & \footnotesize Frame 60 & \footnotesize Frame 80 \\
    \flrowlab{Full\\Attention} &
    \includegraphics[width=0.18\textwidth]{Figures/results_supp/origin_#1_0.jpg} &
    \includegraphics[width=0.18\textwidth]{Figures/results_supp/origin_#1_20.jpg} &
    \includegraphics[width=0.18\textwidth]{Figures/results_supp/origin_#1_40.jpg} &
    \includegraphics[width=0.18\textwidth]{Figures/results_supp/origin_#1_60.jpg} &
    \includegraphics[width=0.18\textwidth]{Figures/results_supp/origin_#1_80.jpg} \\
    \flrowlab{HeadCast} &
    \includegraphics[width=0.18\textwidth]{Figures/results_supp/head_#1_0.jpg} &
    \includegraphics[width=0.18\textwidth]{Figures/results_supp/head_#1_20.jpg} &
    \includegraphics[width=0.18\textwidth]{Figures/results_supp/head_#1_40.jpg} &
    \includegraphics[width=0.18\textwidth]{Figures/results_supp/head_#1_60.jpg} &
    \includegraphics[width=0.18\textwidth]{Figures/results_supp/head_#1_80.jpg} \\}

\begin{figure*}[p]
    \centering
    \begin{subfigure}{\textwidth}
        \centering
        \setlength{\tabcolsep}{1.5pt}
        \renewcommand{\arraystretch}{0.5}
        \begin{tabular}{@{}c@{\hskip 2pt}ccccc@{}}
            \resrowsS{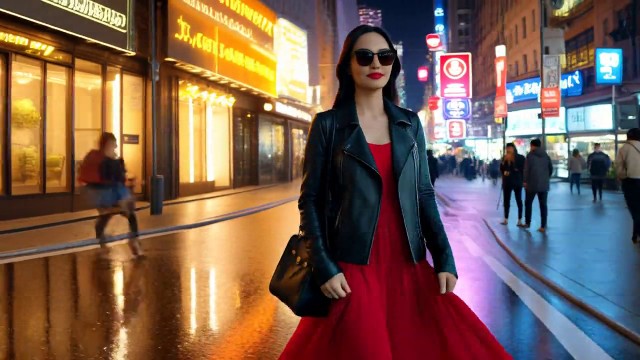}
        \end{tabular}
        \caption{A stylish woman walking down a neon-lit Tokyo street.}
        \label{fig:results_sf_0}
    \end{subfigure}

    \vspace{8pt}

    \begin{subfigure}{\textwidth}
        \centering
        \setlength{\tabcolsep}{1.5pt}
        \renewcommand{\arraystretch}{0.5}
        \begin{tabular}{@{}c@{\hskip 2pt}ccccc@{}}
            \resrowsS{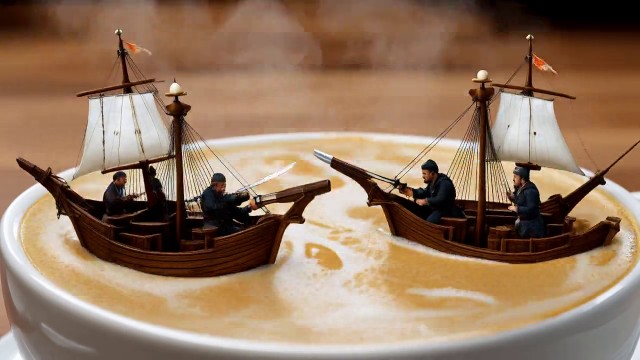}
        \end{tabular}
        \caption{Two pirate ships battling inside a cup of coffee.}
        \label{fig:results_sf_7}
    \end{subfigure}
    \caption{\textbf{Additional qualitative results on Self-Forcing 5s (720P), part~1.} Rows top to bottom: Full Attention, HeadCast. Five frames are sampled uniformly across the 5-second clip.}
    \label{fig:results_supp_sf_a}
\end{figure*}

\begin{figure*}[p]
    \centering
    \begin{subfigure}{\textwidth}
        \centering
        \setlength{\tabcolsep}{1.5pt}
        \renewcommand{\arraystretch}{0.5}
        \begin{tabular}{@{}c@{\hskip 2pt}ccccc@{}}
            \resrowsS{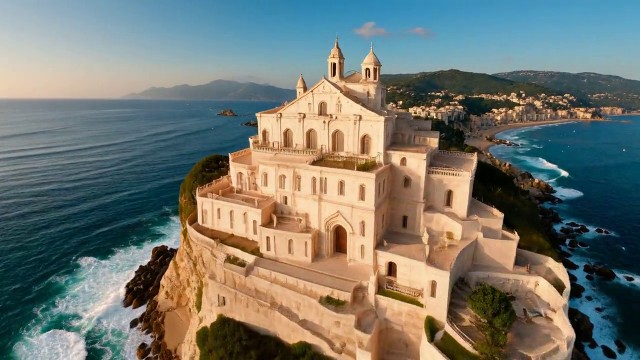}
        \end{tabular}
        \caption{A drone shot circling a historic church on the Amalfi Coast.}
        \label{fig:results_sf_19}
    \end{subfigure}

    \vspace{8pt}

    \begin{subfigure}{\textwidth}
        \centering
        \setlength{\tabcolsep}{1.5pt}
        \renewcommand{\arraystretch}{0.5}
        \begin{tabular}{@{}c@{\hskip 2pt}ccccc@{}}
            \resrowsS{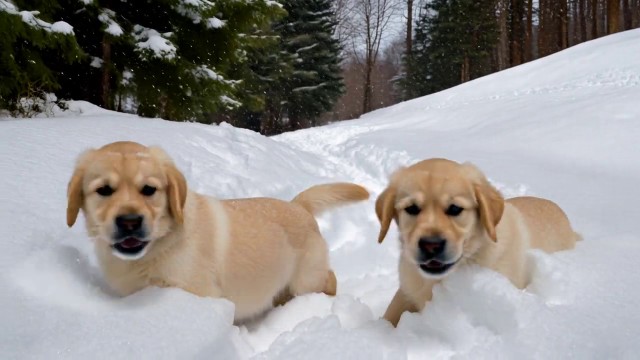}
        \end{tabular}
        \caption{Golden retriever puppies playing in the snow.}
        \label{fig:results_sf_32}
    \end{subfigure}
    \caption{\textbf{Additional qualitative results on Self-Forcing 5s (720P), part~2.} Rows top to bottom: Full Attention, HeadCast. Five frames are sampled uniformly across the 5-second clip.}
    \label{fig:results_supp_sf_b}
\end{figure*}

The longer $30$-second LongLive setting (720P; Figures~\ref{fig:results_supp_a} and~\ref{fig:results_supp_b}) is a harder test, yet HeadCast still matches the baseline across the whole rollout.

\newcommand{\resrows}[1]{%
    & \footnotesize Frame 0 & \footnotesize Frame 125 & \footnotesize Frame 250 & \footnotesize Frame 375 & \footnotesize Frame 500 \\
    \flrowlab{Full\\Attention} &
    \includegraphics[width=0.18\textwidth]{Figures/results_supp/origin_#1_0.jpg} &
    \includegraphics[width=0.18\textwidth]{Figures/results_supp/origin_#1_125.jpg} &
    \includegraphics[width=0.18\textwidth]{Figures/results_supp/origin_#1_250.jpg} &
    \includegraphics[width=0.18\textwidth]{Figures/results_supp/origin_#1_375.jpg} &
    \includegraphics[width=0.18\textwidth]{Figures/results_supp/origin_#1_500.jpg} \\
    \flrowlab{HeadCast} &
    \includegraphics[width=0.18\textwidth]{Figures/results_supp/head_#1_0.jpg} &
    \includegraphics[width=0.18\textwidth]{Figures/results_supp/head_#1_125.jpg} &
    \includegraphics[width=0.18\textwidth]{Figures/results_supp/head_#1_250.jpg} &
    \includegraphics[width=0.18\textwidth]{Figures/results_supp/head_#1_375.jpg} &
    \includegraphics[width=0.18\textwidth]{Figures/results_supp/head_#1_500.jpg} \\}

\begin{figure*}[p]
    \centering
    \begin{subfigure}{\textwidth}
        \centering
        \setlength{\tabcolsep}{1.5pt}
        \renewcommand{\arraystretch}{0.5}
        \begin{tabular}{@{}c@{\hskip 2pt}ccccc@{}}
            \resrows{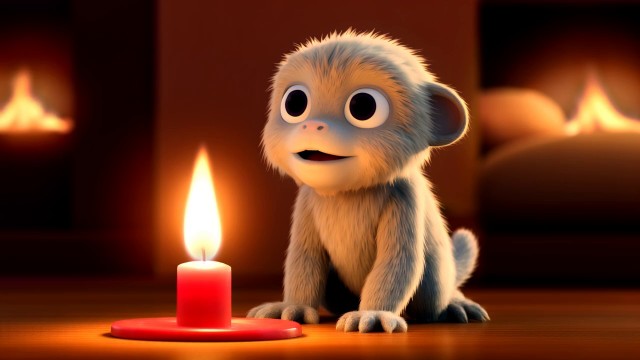}
        \end{tabular}
        \caption{A meerkat figurine beside a lit candle.}
        \label{fig:results_4}
    \end{subfigure}

    \vspace{8pt}

    \begin{subfigure}{\textwidth}
        \centering
        \setlength{\tabcolsep}{1.5pt}
        \renewcommand{\arraystretch}{0.5}
        \begin{tabular}{@{}c@{\hskip 2pt}ccccc@{}}
            \resrows{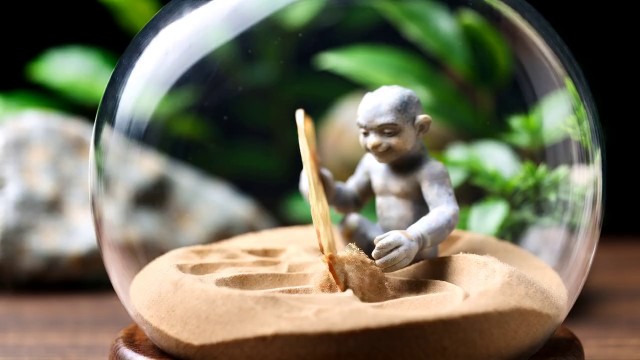}
        \end{tabular}
        \caption{A miniature figure inside a glass dome.}
        \label{fig:results_10}
    \end{subfigure}
    \caption{\textbf{Additional qualitative results on LongLive 30s (720P), part~1.} Rows top to bottom: Full Attention, HeadCast. Five frames are sampled uniformly across the 30-second clip.}
    \label{fig:results_supp_a}
\end{figure*}

\begin{figure*}[p]
    \centering
    \begin{subfigure}{\textwidth}
        \centering
        \setlength{\tabcolsep}{1.5pt}
        \renewcommand{\arraystretch}{0.5}
        \begin{tabular}{@{}c@{\hskip 2pt}ccccc@{}}
            \resrows{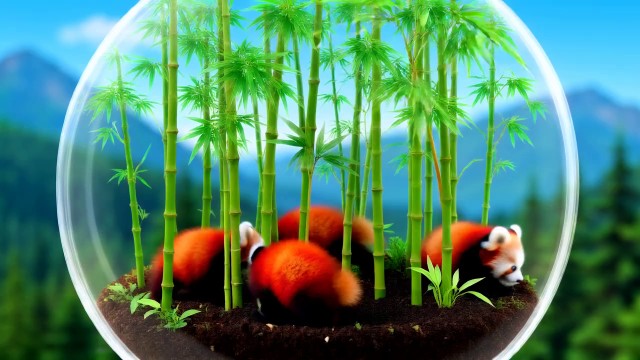}
        \end{tabular}
        \caption{Red pandas in a bamboo terrarium.}
        \label{fig:results_14}
    \end{subfigure}

    \vspace{8pt}

    \begin{subfigure}{\textwidth}
        \centering
        \setlength{\tabcolsep}{1.5pt}
        \renewcommand{\arraystretch}{0.5}
        \begin{tabular}{@{}c@{\hskip 2pt}ccccc@{}}
            \resrows{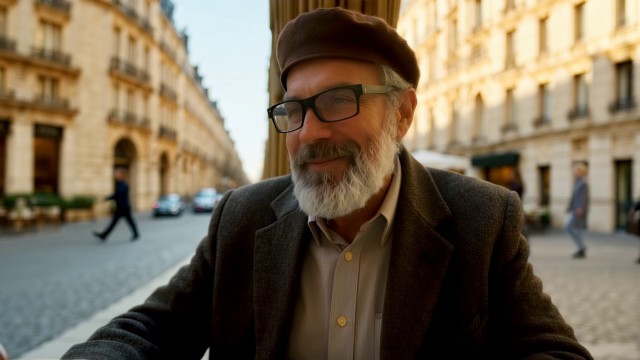}
        \end{tabular}
        \caption{An elderly man walking along a city street.}
        \label{fig:results_29}
    \end{subfigure}
    \caption{\textbf{Additional qualitative results on LongLive 30s (720P), part~2.} Rows top to bottom: Full Attention, HeadCast. Five frames are sampled uniformly across the 30-second clip.}
    \label{fig:results_supp_b}
\end{figure*}

\section{User Study Protocol}
\label{sec:user_study_protocol}

To complement the automatic metrics with human judgment, we run a user study based on the Two-Alternative Forced Choice (2AFC) protocol. In each question a participant is shown two videos generated from the \emph{same} prompt and answers a single question---\emph{which video do you prefer overall?}---picking the one that looks more realistic and natural, with fewer visible artifacts. We deliberately collect one holistic preference rather than splitting the judgment across sub-dimensions: the artifacts our method targets---inter-frame flicker and structural drift at block boundaries---degrade the overall viewing experience rather than any single attribute, so a forced overall choice is the most direct measure of which video a viewer would rather watch.

We evaluate HeadCast in two head-to-head settings:
\begin{itemize}
    \item \textbf{HeadCast vs.\ Full Attention}, which tests whether our acceleration preserves the quality of the unmodified model; and
    \item \textbf{HeadCast vs.\ Dummy Forcing}, the most directly comparable training-free baseline, which tests whether our head-specific eviction yields visibly better video than aggressive eviction.
\end{itemize}

For each setting we generate $160$ same-prompt pairs spanning both clip lengths---$128$ short ($5$\,s) and $32$ long ($30$\,s)---with each pair sharing prompt and seed so the only difference is the attention pathway. The prompts are sampled uniformly at random from the Movie Gen prompt suite. We then split the $160$ pairs into $10$ groups by randomly assigning pairs within each length stratum, so the grouping is unbiased while every group still holds a comparable mix ($\sim$$12$--$13$ short and $3$--$4$ long clips). Every group is scored by two \emph{different} annotators in two independent rounds, giving two votes per pair ($\sim$$320$ per setting) over $20$ participants in total.

Within each pair the two videos are shown as \emph{video~1}/\emph{video~2} in randomized order, with the slot-to-method mapping withheld and stored separately, so the study is fully double-blind; each question is a strict two-way forced choice with no ``tie'' option. We report the percentage of votes preferring HeadCast over its opponent (above $50\%$ favors HeadCast); results are in Section~\ref{sec:user_study}.

\end{document}